\documentclass[pdflatex,sn-nature]{sn-jnl}

\usepackage{graphicx}%
\usepackage{multirow}%
\usepackage{amsmath,amssymb,amsfonts}%
\usepackage{amsthm}%
\usepackage{mathrsfs}%
\usepackage[title]{appendix}%
\usepackage{xcolor}%
\usepackage{textcomp}%
\usepackage{manyfoot}%
\usepackage{booktabs}%
\usepackage{algorithm}%
\usepackage{algorithmicx}%
\usepackage{algpseudocode}%
\usepackage{listings}%
\usepackage{array}
\usepackage{xspace}
\usepackage{xurl}
\usepackage[resetlabels]{multibib}
\newcites{app}{References}

\newcommand{\eg}{\emph{e.g.,}\xspace}

\newcommand{\etal}{\emph{et al.,}\xspace}
\usepackage{caption}
\usepackage{color}
\usepackage[most]{tcolorbox}
\usepackage{bbding}
\usepackage{pifont}
\usepackage{makecell}
\usepackage{colortbl}
\usepackage{subcaption,siunitx,booktabs}
\usepackage[defaultcolor=magenta,commandnameprefix=always]{changes}
\usepackage{adjustbox}
\usepackage{longtable}
\usepackage{tabularx}
\usepackage{ltxtable}
\usepackage{xltabular}
\usepackage{lineno}

\theoremstyle{thmstyleone}%
\theoremstyle{thmstyletwo}%

\theoremstyle{thmstylethree}%

\begin{document}

\title[Article Title]{Conversational Interaction for LLM-Based Clinical Text-to-SQL: A User Study with Healthcare Professionals}


\author[1]{\fnm{Jun-Min} \sur{Lee}}\email{ljm565@kaist.ac.kr}
\equalcont{These authors contributed equally to this work.}

\author[2]{\fnm{Suyoung} \sur{Yoo}}\email{rueseo92@gmail.com}
\equalcont{These authors contributed equally to this work.}

\author[1]{\fnm{Gyubok} \sur{Lee}}\email{gyubok.lee@kaist.ac.kr}

\author*[1]{\fnm{Edward} \sur{Choi}}\email{edwardchoi@kaist.ac.kr}

\author*[3,4]{\fnm{Meong Hi} \sur{Son}}\email{meonghison@gmail.com}

\affil[1]{\orgdiv{Department of Kim Jaechul Graduate School of AI}, \orgname{Korea Advanced Institute of Science and Technology}, \orgaddress{\city{Seongnam-si}, \postcode{13558}, \state{Gyeonggi-do}, \country{Republic of Korea}}}

\affil[2]{\orgdiv{Research Institute for Future Medicine}, \orgname{Samsung Medical Center}, \orgaddress{\city{Seoul}, \postcode{06351}, \country{Republic of Korea}}}

\affil[3]{\orgdiv{Department of Emergency Medicine}, \orgname{Samsung Medical Center}, \orgaddress{\city{Seoul}, \postcode{06351}, \country{Republic of Korea}}}

\affil[4]{\orgdiv{Office of AI Transformation}, \orgname{Samsung Medical Center}, \orgaddress{\city{Seoul}, \postcode{06351}, \country{Republic of Korea}}}


\abstract{
Conversational text-to-SQL interfaces could give healthcare professionals natural-language access to electronic health record (EHR) data, but evidence for their real-world usability remains limited: prior evaluations have relied largely on curated benchmarks with fully specified questions or on simulated users.
We report a user study in which 14 physicians who regularly work with EHR data used both a single-turn and a conversational multi-turn text-to-SQL interface over the MIMIC-IV database in a within-subjects design; dialogue analyses additionally drew on interaction logs from 21 healthcare professionals across the main study and two pilot studies.
The conversational interface was consistently favored: mean System Usability Scale scores rose from 68.2, near the industry average, to 85.9 ($p<0.001$), with significant advantages across technology-acceptance constructs including performance expectancy, effort expectancy, self-efficacy, and behavioral intention.
In post-study forced-choice comparisons, 93\% of participants reported greater overall satisfaction with the conversational interface, 86\% preferred it for real-world adoption, and 71\% trusted its results more; no participant preferred or trusted the single-turn interface more.
Turn-level analysis of 425 logged user turns showed that freely composed queries triggered clarification requests in 24.8\% of cases versus 1.8\% for predefined, fully specified questions, and 64.3\% of clarification episodes stemmed from EHR- or clinical-domain knowledge.
Conversational interaction is a practical prerequisite, rather than a convenience, for natural-language EHR access. 
The frequent need for clarification in realistic clinical queries further underscores the importance of conversational interfaces, while benchmark-style evaluation can substantially underrepresent the ambiguity of real clinical use.
}

\keywords{text-to-SQL, electronic health records (EHR), conversational agents, large language models (LLM), usability evaluation, human-AI interaction}

\maketitle

\section{Introduction}
\label{sec:introduction}
Electronic Health Records (EHRs) hold decades of longitudinal clinical data with substantial value for clinical research, quality improvement, and decision support~\cite{charles2015adoption,jensen2012mining}.
In practice, however, these data remain difficult to access: they are distributed across complex relational schemas with numerous interconnected tables, heterogeneous structures, and incompletely standardized terminology~\cite{holmes2021electronic,miller2024automated}.
Extracting a patient cohort can require both Structured Query Language (SQL) proficiency and detailed knowledge of the underlying database, creating barriers for healthcare professionals without database expertise~\cite{holmes2021electronic,olakotan2025usability}.
Natural-language interfaces could reduce these barriers by allowing users to express their information needs without writing SQL directly.

As Large Language Models (LLMs) have rapidly advanced in natural language understanding, reasoning, and code generation, interest in text-to-SQL systems that translate natural language questions into executable SQL queries has grown substantially.
Building on early neural semantic parsers and large-scale benchmarks~\cite{zhong2017seq2sql,yu2018spider}, recent LLM-based systems approach expert-level accuracy on complex database schemas~\cite{pourreza2023dinsql,hong2024nextgeneration}.
Alongside these advances, healthcare text-to-SQL systems have evolved from single-turn querying~\cite{wang2020text,lee2022ehrsql} toward conversational multi-turn interaction~\cite{ryu2024ehr}, raising the prospect of natural language access to EHR data for users without database expertise~\cite{ohno2011realizing,yang2022large}.
Despite these advances, evaluation has remained predominantly benchmark-based, with an emphasis on SQL generation and execution accuracy~\cite{lee2022ehrsql,3666122.3667957}.
How healthcare professionals actually experience such interfaces---whether they can express their information needs, resolve misunderstandings, trust the returned answers, and would adopt these tools in practice---remains largely unexamined, leaving a gap between benchmark performance and real-world usability~\cite{ipeirotis2025nlidb}.

This gap matters because questions posed in real use may differ substantially from the fully specified questions found in text-to-SQL benchmarks.
Naturally expressed information needs are frequently ambiguous~\cite{min2020ambigqa,wang-etal-2023-know}, making clarification an important mechanism for recovering the user's true intent~\cite{rao2018learning,yao2019model}.
However, existing medical text-to-SQL studies have largely evaluated systems on fully specified queries with ambiguity removed in advance~\cite{wang2020text,lee2022ehrsql,ryu2024ehr}.
The recently proposed EHR-ChatQA benchmark~\cite{lee2026conversationqueryexecutionbenchmarking} takes a step toward closing this gap by evaluating agents that resolve ambiguity through dialogue, but it relies on LLM-simulated users.
It therefore remains unknown how often clarification is needed when healthcare professionals formulate their own EHR questions, what triggers it, and whether conversational interaction translates into greater usability and trust in practice.

Here we report, to our knowledge, the first user study evaluating conversational text-to-SQL for EHR data with real healthcare professionals rather than simulated users.
In the main study, 14 physicians who regularly work with EHR data used both a single-turn and a conversational multi-turn interface over the MIMIC-IV database~\cite{johnson2023mimic} in a within-subjects design.
For the dialogue analysis, we additionally analyzed all interaction logs from the main study and two pilot studies, comprising dialogues from 21 healthcare professionals, including physicians, nurses, and clinical data-team members.
The conversational interface was consistently favored: mean System Usability Scale (SUS)~\cite{brooke1996sus} scores increased from 68.2 to 85.9, with significant advantages across multiple Unified Theory of Acceptance and Use of Technology (UTAUT)~\cite{venkatesh2003user} constructs, and no participant preferred or trusted the single-turn interface more.
Turn-level analysis further showed that freely composed clinician queries triggered clarification in 24.8\% of cases, compared with 1.8\% for predefined, fully specified questions, and that 64.3\% of clarification episodes arose from EHR- or clinical-domain knowledge.
Together, these findings suggest that conversational interaction is not merely a usability convenience but a core capability for natural language access to EHR data.
They further indicate that benchmark-style evaluation can substantially underrepresent the ambiguity encountered in real use, and that the identified sources of clarification provide concrete targets for the design and evaluation of future clinical text-to-SQL systems.


\section{Results}
\label{sec:results}

\begin{figure}[t!]
    \centering
    \includegraphics[width=\textwidth]{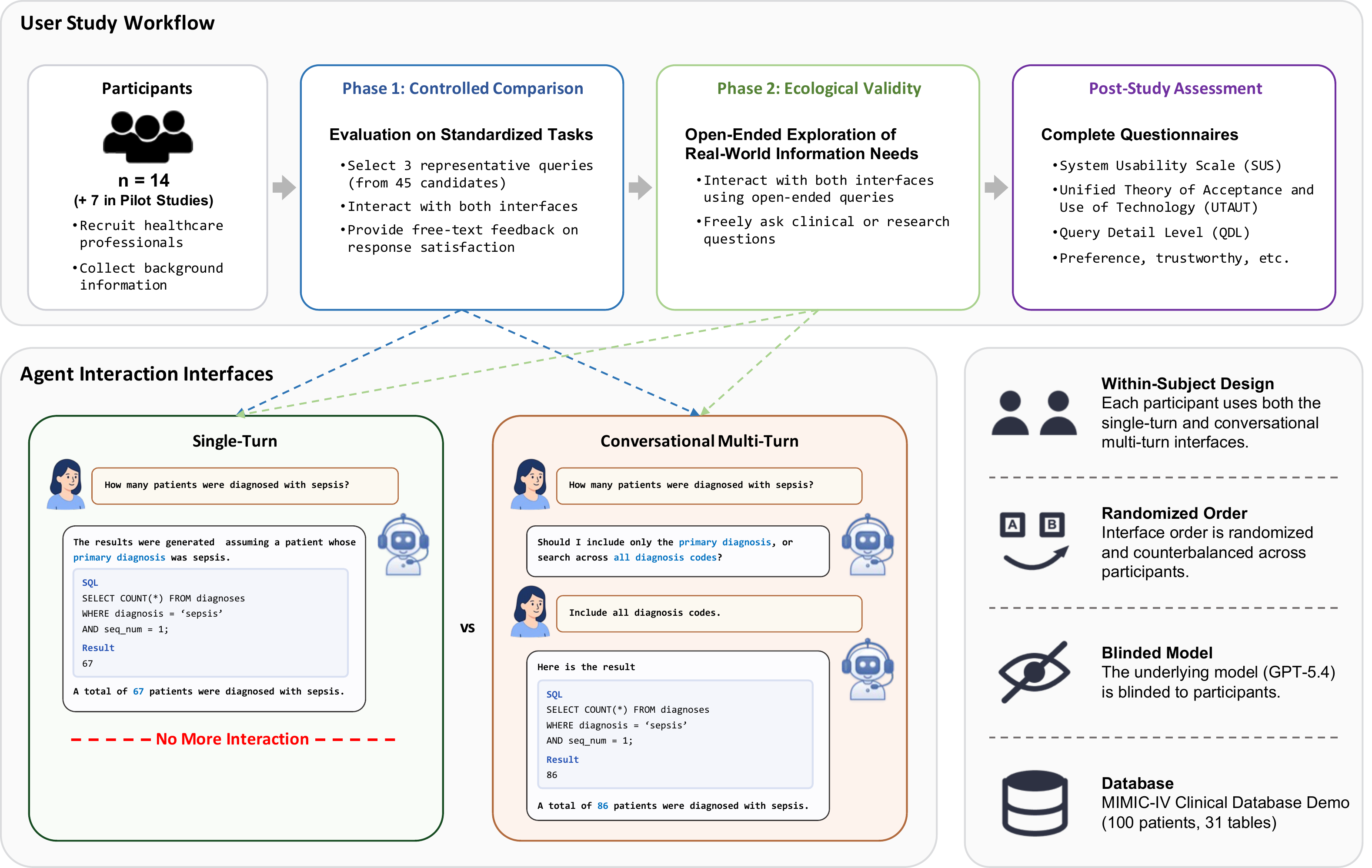}
    \caption{Overview of the user-study design. Participants completed a within-subject comparison of single-turn and conversational multi-turn text-to-SQL interfaces, consisting of a controlled comparison on pre-defined questions (Phase 1) followed by open-ended querying (Phase 2).}
    \label{fig:design}
\end{figure}

We evaluated the single-turn and conversational multi-turn text-to-SQL interfaces in a within-subjects study involving 14 physicians.
The study consisted of a controlled comparison using predefined questions (Phase 1), followed by open-ended querying intended to capture more natural information-seeking behavior (Phase 2) (Fig.~\ref{fig:design}).
Section~\ref{sec:results:participant_characteristics} describes the study cohort and participant characteristics.
Section~\ref{sec:results:usability_evaluation} reports usability, technology acceptance, preference, and trust outcomes.
Section~\ref{sec:results:real_dialogue_analysis} examines user queries and conversational behavior observed during the study.

\subsection{Participant Characteristics}
\label{sec:results:participant_characteristics}

\begin{figure}[t!]
\centering
\includegraphics[width=\textwidth]{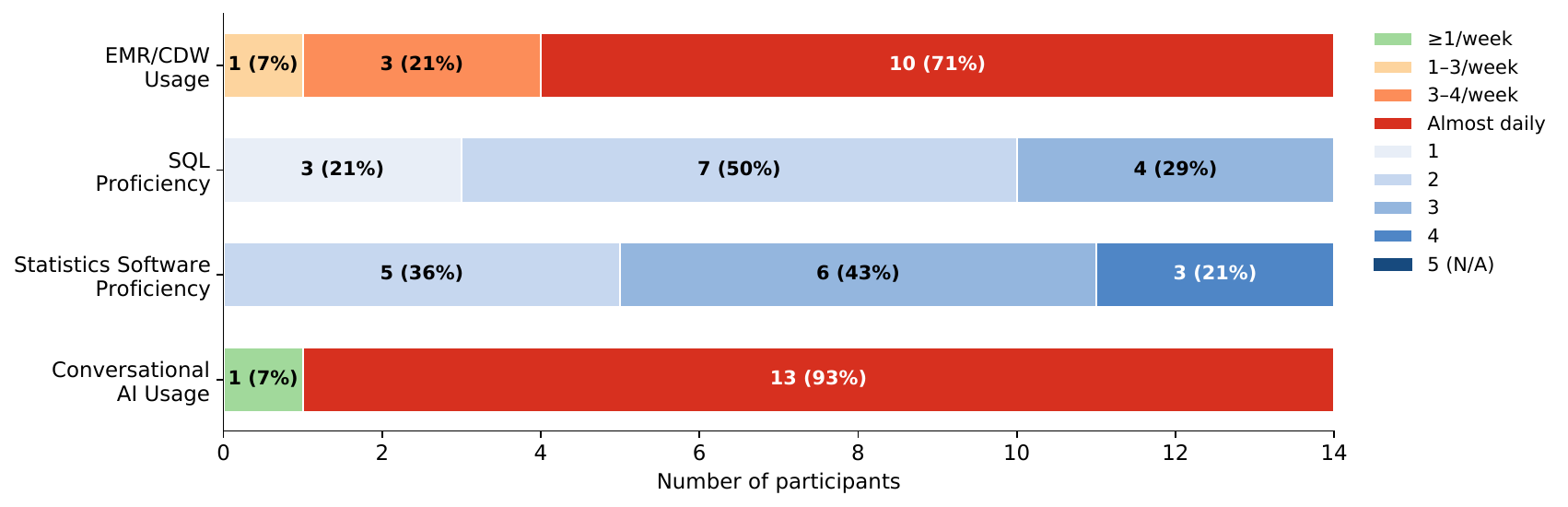}
    \caption{Distribution of participant background characteristics ($n=14$), including EHR (or EMR)/CDW usage frequency, SQL proficiency, statistical software proficiency, and conversational AI usage frequency.}
\label{fig:participant_background}
\end{figure}

Participant background characteristics are summarized in Fig.~\ref{fig:participant_background}.
Most participants reported near-daily use of EHR (or EMR)/CDW\footnote{EHR: Electronic Health Record; EMR: Electronic Medical Record; CDW: Clinical Data Warehouse} systems, while SQL proficiency was generally low and proficiency with statistical analysis tools was modest.
Conversational AI familiarity was high, with nearly all participants reporting almost daily use.
Overall, the cohort consisted of physicians with substantial EHR experience but limited expertise in direct database querying.

\subsection{Usability Evaluation}
\label{sec:results:usability_evaluation}
After completing both study phases, participants evaluated each interface using SUS and UTAUT constructs.
Higher SUS scores indicate greater perceived usability.
The UTAUT evaluation included Performance Expectancy (PE), Effort Expectancy (EE), Self-Efficacy (SE), Anxiety (ANX), and Behavioral Intention (BI), with higher scores indicating more favorable perceptions except for ANX.
We additionally report Avg.~UTAUT (P), the average of the four positively oriented constructs (PE, EE, SE, and BI).
Qualitative usability was examined through open-ended survey responses.

\subsubsection{Usability and Technology Acceptance}
\label{sec:results:usability_evaluation:inferential_statistical_results}

\begin{figure}[t!]
    \centering
    \includegraphics[width=\textwidth]{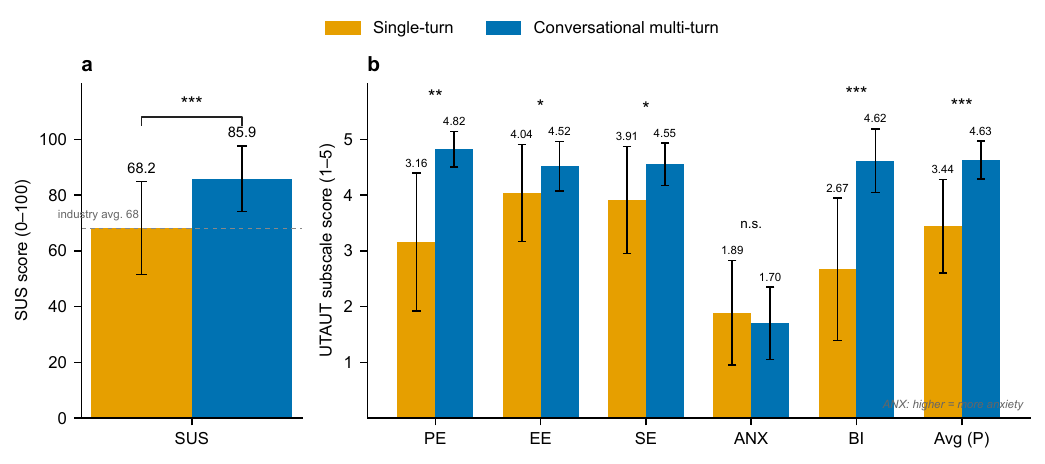}\par\vspace{6pt}
    \begin{minipage}{\textwidth}
        \raggedright{\sffamily\bfseries c}\par\vspace{2pt}
        \centering
        \includegraphics[width=0.92\textwidth]{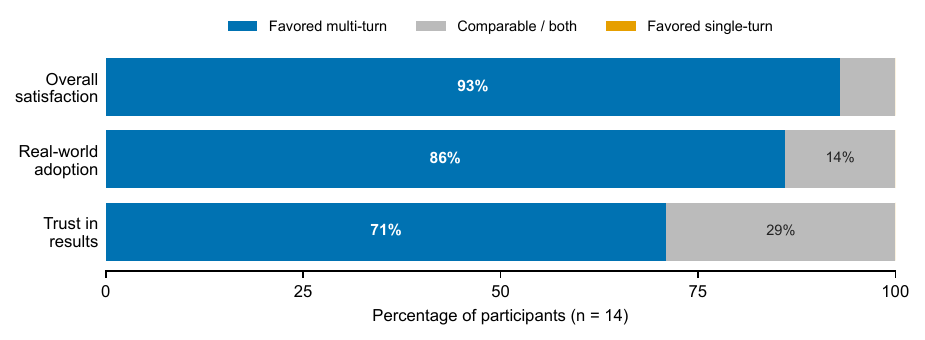}
    \end{minipage}
    \caption{System usability, technology acceptance, and interface preference for the single-turn versus conversational multi-turn interfaces ($n=14$). (a) SUS scores (0--100); the dashed line indicates the industry-average score of approximately 68. (b) UTAUT construct scores (1--5): Performance Expectancy (PE), Effort Expectancy (EE), Self-Efficacy (SE), Anxiety (ANX), Behavioral Intention (BI), and the average of the four positively oriented constructs, Avg.\ UTAUT (P). Bars represent the mean and error bars indicate the standard deviation. Statistical significance is based on the primary paired test for each metric (Section~\ref{sec:method:usability_analysis:statistical_comparison_procedure}): $^{*}p<0.05$, $^{**}p<0.01$, $^{***}p<0.001$; n.s., not significant. Higher ANX scores indicate greater anxiety. (c) Distribution of participant responses to the three post-study forced-choice comparison questions ($n=14$): overall satisfaction, preferred interface for real-world adoption, and trust in the returned results.}
    \label{fig:usability}
\end{figure}

Fig.~\ref{fig:usability}a,b summarizes the paired comparisons of SUS and UTAUT outcomes between the two interfaces ($n=14$); complete descriptive and inferential statistics are provided in Appendix~\ref{app:supplementary_results:supplementary_statistical_results}.
Wilcoxon signed-rank tests were used as the primary tests for PE, EE, and SE because paired differences deviated from normality; paired $t$-tests were used for the remaining metrics.

The mean SUS score increased from 68.2 for the single-turn interface to 85.9 for the conversational multi-turn interface, corresponding to \textit{Good} and \textit{Best Imaginable}, respectively, on the adjective-rating scale proposed by Bangor \etal~\cite{bangor2009determining}.
SUS, PE, EE, SE, BI, and Avg.~UTAUT (P) were significantly higher for the conversational multi-turn interface than for the single-turn interface ($p<0.05$), whereas ANX showed no statistically significant difference.
Effect sizes were consistent with the inferential results.
Cohen's $d_z$ indicated large effects for SUS, PE, BI, and Avg.~UTAUT (P) ($|d_z|>1.3$) and medium-to-large effects for EE ($|d_z|=0.75$) and SE ($|d_z|=0.72$); Wilcoxon effect sizes were large for all statistically significant metrics ($|r|\geq0.5$)~\cite{brydges2019effect}.

\subsubsection{Associations with Participant Characteristics}
\label{sec:results:usability_evaluation:participant_characteristics}
To examine whether the usability benefits of the conversational multi-turn interface varied according to participants' query formulation styles or technical proficiency, we assessed the associations of SUS-based usability outcomes with Query Detail Level (QDL) and self-reported proficiency measures.
QDL was assessed by presenting participants with three clinical information-seeking scenarios and asking them to rate, on a 1-to-5 scale anchored by example queries at each end, how they would typically formulate a query for each scenario. 
QDL ratings varied substantially across participants and clinical information-seeking scenarios (Appendix~\ref{app:qdl_distribution}), and Spearman correlation analysis showed no significant association between participant-level mean QDL scores and SUS scores for either the single-turn ($\rho=-0.12$, $p=0.68$) or conversational multi-turn interface ($\rho=-0.05$, $p=0.87$). 
Conversational-interface usability gains, defined as $\mathrm{SUS}_{\mathrm{Multi\text{-}turn}}-\mathrm{SUS}_{\mathrm{Single\text{-}turn}}$, were also not significantly associated with SQL proficiency ($\rho=-0.29$, $p=0.32$) or statistical software proficiency ($\rho=-0.32$, $p=0.26$).

\subsubsection{Overall Preference and Trust}
\label{sec:results:usability_evaluation:preference_trust}

Fig.~\ref{fig:usability}c summarizes responses to the three post-study forced-choice comparisons.
For overall satisfaction, 13 of 14 participants (93\%) rated the conversational multi-turn interface as ``much'' more satisfying, and the remaining participant (7\%) rated it as ``somewhat'' more satisfying.
For real-world adoption, 12 participants (86\%) preferred the conversational multi-turn interface, while 2 (14\%) indicated that they would use both interfaces depending on the task.
For trust in the returned results, 10 participants (71\%) reported greater trust in the conversational multi-turn interface, while 4 (29\%) reported comparable trust in the two interfaces.
No participant preferred or trusted the single-turn interface over the conversational multi-turn interface.

\subsection{Real Dialogue Analysis}
\label{sec:results:real_dialogue_analysis}
Real dialogues collected from the main study ($n=14$) and two pilot studies ($n=7$) were analyzed at the turn level.
In total, 183 dialogue sessions were annotated using seven turn types: four user-turn types (Q, FX-Q, SF-Q, and CL-Q) and three agent-response types (ANS, S-ANS, and C-ANS).
Clarification interactions (CL-Q/C-ANS) and follow-up suggestion interactions (SF-Q/S-ANS) occurred only in the conversational multi-turn interface.
Detailed turn definitions and the distribution of turn types by interface are provided in Appendix~\ref{app:supplementary_results:turn_definition}.

\subsubsection{Clarification and Follow-up Interactions}
\label{sec:results:real_dialogue_analysis:clarification}
\begin{table*}[t!]
    \caption{Clarification occurrence and resolution in the conversational multi-turn interface, and user acceptance of agent-proposed follow-up suggestions.}
    \label{tab:clarification_summary}
    \centering
    \resizebox{1\linewidth}{!}{
    \begin{tabular}{l l r r}
    \toprule

    \multicolumn{4}{l}{\textbf{Occurrence of Clarification Requests (C-ANS)}} \\
    \midrule
    \textbf{Preceding Turn Type} & \textbf{Turn Sequence} & \textbf{C-ANS / Total} & \textbf{Rate} \\
    \midrule
    Question (Q)    & Q $\rightarrow$ C-ANS   & 40 / 161 & 24.8\% \\
    Fixed Question (FX-Q)  & FX-Q $\rightarrow$ C-ANS & 1 / 56 & 1.8\% \\
    Suggested Follow-up Question (SF-Q)  & SF-Q $\rightarrow$ C-ANS & 1 / 60 & 1.7\% \\
    Clarification Response (CL-Q)  & CL-Q $\rightarrow$ C-ANS & 6 / 46 & 13.0\% \\
    \addlinespace
    \textbf{Total} & & \textbf{48 / 323} & \textbf{14.9\%} \\

    \midrule

    \multicolumn{4}{l}{\textbf{Resolution of Clarification Exchanges}} \\
    \midrule
    \textbf{Required Rounds} & \textbf{Turn Sequence} & \textbf{Exchanges (Turns)} & \textbf{Proportion} \\
    \midrule
    1 round &
    $\rightarrow$ C-ANS $\rightarrow$ CL-Q &
    36 (36) &
    85.7\% \\

    2 rounds &
    $\rightarrow$ C-ANS $\rightarrow$ CL-Q
    $\rightarrow$ C-ANS $\rightarrow$ CL-Q &
    2 (4) &
    4.8\% \\

    3 rounds &
    $\rightarrow$ C-ANS $\rightarrow$ CL-Q
    $\rightarrow$ C-ANS $\rightarrow$ CL-Q
    $\rightarrow$ C-ANS $\rightarrow$ CL-Q &
    2 (6) &
    4.8\% \\

    Unresolved &
    $\rightarrow$ C-ANS $\rightarrow$ \textit{(no further CL-Q recorded)} &
    2 (2) &
    4.8\% \\
    \addlinespace
    \textbf{Total} & & \textbf{42 (48)} & \textbf{100\%} \\

    \midrule
    \midrule

    \multicolumn{4}{l}{\textbf{User Acceptance of Follow-up Suggestions (S-ANS)}} \\
    \midrule
    \multicolumn{2}{l}{\textbf{Category}} & \multicolumn{2}{r}{\textbf{Turns}} \\
    \midrule
    \multicolumn{2}{l}{All S-ANS turns} & \multicolumn{2}{r}{249} \\
    \multicolumn{2}{l}{\quad Excluded: Session-final S-ANS turns (unobservable)} & \multicolumn{2}{r}{66} \\
    \midrule
    \multicolumn{2}{l}{Eligible S-ANS turns} & \multicolumn{2}{r}{183 ($=249 - 66$)} \\
    \multicolumn{2}{l}{\quad Accepted (followed by an SF-Q)} & \multicolumn{2}{r}{60} \\
    \midrule
    \multicolumn{2}{l}{\textbf{Acceptance rate}} & \multicolumn{2}{r}{\textbf{32.8\% (60/183)}} \\

    \bottomrule
    \end{tabular}
    }
\end{table*}
A clarification request was identified when a user turn was immediately followed by an agent response labeled C-ANS.
As shown in the upper section of Table~\ref{tab:clarification_summary}, fixed questions (FX-Q) and agent-suggested follow-up questions (SF-Q) rarely triggered clarification requests (1.8\% and 1.7\%, respectively), whereas freely formulated user questions (Q) triggered clarification in 24.8\% of cases.
Additional clarification was requested after 13.0\% of CL-Q turns, in which the user had responded to the preceding agent clarification request (C-ANS $\rightarrow$ \textbf{CL-Q $\rightarrow$ C-ANS}).
Among 42 clarification exchanges, 40 were successfully resolved; of these, 36 were resolved within a single clarification round, while four required multiple rounds.
The remaining two were unresolved because the clarification request occurred at the end of the session, with no subsequent participant response.

The agent generated 249 follow-up suggestions (S-ANS). 
Of these, 183 were eligible for acceptance analysis. 
The remaining suggestions were excluded because they were presented in the final agent response of the dialogue session, leaving no subsequent user turn in which acceptance or rejection could be observed. 
Among the 183 eligible suggestions, 60 were accepted in the next user turn, yielding an acceptance rate of 32.8\%, as shown in the lower section of Table~\ref{tab:clarification_summary}.

\subsubsection{Typology of Clarification Triggers}
\label{sec:results:real_dialogue_analysis:typology}
\begin{figure}[t!]
\centering
\includegraphics[width=0.7\textwidth]{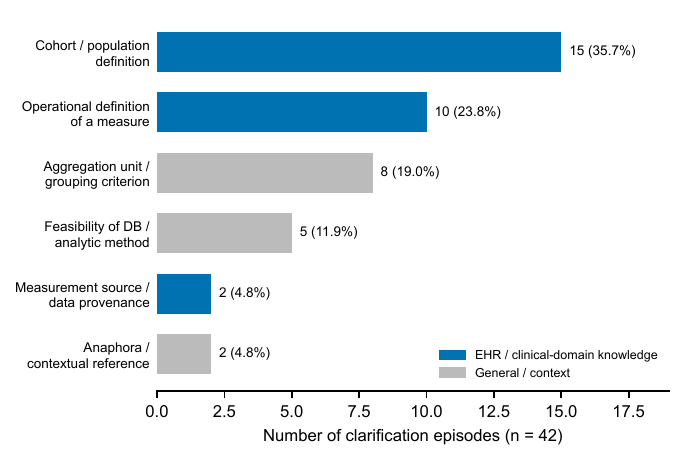}
    \caption{Distribution of the 42 clarification exchanges by dominant trigger category. Categories shaded in blue indicate triggers requiring EHR- or clinical-domain knowledge.}
\label{fig:clarification}
\end{figure}
The 42 clarification exchanges were qualitatively categorized according to the dominant trigger of the clarification request.
Each exchange was assigned to one of six mutually exclusive categories; for multi-round exchanges, only the initial C-ANS turn was coded.
Detailed coding criteria and representative examples are provided in Appendix~\ref{app:supplementary_results:clarification_typology}.

As shown in Fig.~\ref{fig:clarification}, \textit{cohort/population definition} was the most frequent trigger (35.7\%), followed by \textit{operational definition of a measure} (23.8\%) and \textit{aggregation unit/grouping criterion} (19.0\%).
The three categories requiring EHR- or clinical-domain knowledge---\textit{cohort/population definition}, \textit{operational definition of a measure}, and \textit{measurement source/data provenance}---accounted for 64.3\% of all clarification exchanges.

\section{Discussion}
\label{sec:Discussion}

The conversational multi-turn interface received significantly higher ratings across all positive usability and technology-acceptance measures and was strongly preferred overall. 
The study cohort consisted of physicians with substantial EHR experience but limited expertise in direct data querying, representing a population in which natural-language EHR access is particularly relevant. 
Within this cohort, differences between the two interfaces were particularly pronounced for PE and BI, whereas those for EE and SE were more modest. 
This pattern suggests that the advantages of conversational interaction lie more in perceived usefulness and intention for future use than in reducing effort or increasing confidence. 
Participants' familiarity with conversational AI (Fig.~\ref{fig:participant_background}) may also have contributed to the relatively modest differences observed in EE and SE. 
ANX was the only measure that did not differ significantly between the two interfaces, and ANX scores were low for both, suggesting that the additional conversational interaction did not increase users' anxiety or psychological burden.

\vspace{0.8em}\noindent
Because the benefit of conversational interaction could vary according to how specifically participants formulate their queries and how familiar they are with data-querying tools, we examined whether users with less detailed query formulation styles or lower technical proficiency derived greater usability benefits from the conversational multi-turn interface. 
Specifically, we assessed associations between QDL and interface-specific SUS scores, as well as between usability gains, defined as $\mathrm{SUS}_{\mathrm{Multi\text{-}turn}}-\mathrm{SUS}_{\mathrm{Single\text{-}turn}}$, and self-reported SQL and statistical software proficiency.
Overall, no significant associations were observed between SUS-based usability outcomes and QDL, SQL proficiency, or statistical software proficiency, suggesting that the usability advantage of conversational interaction was not confined to users with less detailed query formulation styles or lower technical proficiency. 
Rather, the higher usability ratings of the multi-turn interface appeared to extend across participants with differing query formulation styles and levels of technical proficiency. 
However, given the modest sample size ($n=14$), smaller but potentially meaningful associations cannot be excluded.

\vspace{0.8em}\noindent
Clarification behavior differed markedly between predefined and naturally formulated queries. Clinician-authored queries triggered clarification requests in 24.8\% of cases, compared with only 1.8\% for predefined, highly specified questions. 
This suggests that naturally formulated clinical queries are often ambiguous, creating a risk that single-turn agents may resolve the ambiguity in ways that do not fully align with the user's true intent.
Furthermore, among the 42 clarification exchanges, 64.3\% involved ambiguities requiring EHR- or clinical-domain knowledge. 
These cases most commonly concerned cohort or population definitions, operational definitions of clinical measures, and the source or provenance of measurements. 
For example, a seemingly simple clinical concept could require clarification about which patients should be included, how a variable should be operationalized, or which of several available EHR data sources should be used. 
Other clarification triggers involved aggregation or grouping choices, feasibility constraints imposed by the available database, and ambiguous references to prior conversational context. 
Together, these findings indicate that clarification in clinical text-to-SQL is not driven solely by underspecified wording, but also by domain-specific decisions about how clinical concepts should be mapped to EHR data.
Explicit, documented guidance on these recurring ambiguity types may therefore help improve the consistency of clarification behavior across queries and users. 
Further details on the clarification-trigger taxonomy, coding criteria, and representative examples are provided in Appendix~\ref{app:supplementary_results:clarification_typology}.

\vspace{0.8em}\noindent
Although follow-up suggestions were frequently cited as a strength of the conversational interface, only 32.8\% of agent-proposed follow-up tasks were accepted in the next user turn. 
In qualitative feedback, participants valued that the system \textit{``suggested likely follow-up questions"} and \textit{``provided examples of additional questions to consider." }
This contrast suggests that participants appreciated the availability of follow-up suggestions, even when individual suggestions did not always align with their immediate information needs.
The suggestions often involved relatively simple next steps, such as modifying cohort criteria or adding query conditions, rather than reflecting the broader clinical or analytical context of the ongoing exploration. 
This indicates that suggestion generation was driven more by locally plausible query transformations than by the individual user's evolving information-seeking context. 
Improving the agent's clinical-context reasoning may therefore increase the relevance and utility of follow-up suggestions.

\vspace{0.8em}\noindent
Open-ended qualitative feedback further highlighted differences in how participants experienced the two interfaces. 
Negative feedback on the single-turn interface centered primarily on the inability to verify or refine results through follow-up questions. 
In contrast, the conversational multi-turn interface was valued for enabling users to progressively refine their questions and verify returned results through follow-up dialogue, particularly for more complex queries, as well as for suggesting relevant follow-up tasks that users could pursue next. 
Participants noted that users may not always be able to formulate a fully specified question at the outset, particularly in clinical settings where workload and time pressure can limit how precisely information needs are expressed. 
Continued dialogue therefore appeared to reduce the burden of specifying all query details upfront by allowing users to clarify and refine their intent incrementally.
Participants also described follow-up interaction as important for verifying and correcting potentially uncertain outputs. 
This suggests that trust was shaped not only by the correctness of the initial response, but also by whether users could interrogate, revise, and reconcile results through subsequent dialogue. 
Despite these advantages, participants also reported concerns related to failures in context preservation, excessive clarification, response reliability and accuracy, and incomplete consideration of clinically relevant variables. 
These observations suggest that effective conversational interaction requires reliable context maintenance, appropriately calibrated clarification, and more comprehensive clinical reasoning. 
At the same time, two participants indicated that they would still prefer the single-turn interface for simple, well-defined questions requiring quick answers, suggesting that the two interaction modes may serve complementary roles depending on task complexity. 
Representative participant responses supporting these observations are provided in Appendix~\ref{app:representative_feedback}.

\vspace{0.8em}\noindent
In this within-subjects study of physicians, the conversational multi-turn text-to-SQL interface showed clear advantages over the single-turn interface in usability, technology acceptance, and overall preference. 
Naturally formulated clinical queries were substantially more likely to prompt clarification than predefined, highly specified questions, particularly when ambiguity involved EHR- or clinical-domain knowledge. 
These findings suggest that the value of conversational interaction extends beyond improving user experience: it also provides a mechanism for resolving ambiguity that may be difficult to address reliably in a single-turn setting. 
Although the modest sample size and use of the MIMIC-IV Demo dataset may limit generalizability to larger and more complex clinical databases, the consistent usability gains and observed clarification behavior support conversational interaction as an important component of practical natural-language EHR access.

\section{Methods}
\label{sec:methods}

\subsection{User Study Design}
\label{sec:methods:user_study_design}
The study recruited healthcare professionals who routinely used EHRs, including physicians, nurses, and clinical data specialists.
To reduce inter-participant variability arising from differences in SQL proficiency, EHR experience, and query formulation behavior, we adopted a within-subjects design in which every participant used both interfaces.
The order of interface exposure was counterbalanced using randomized assignment to mitigate order effects.

\subsection{Data and System Setup}
\label{sec:methods:data_and_system_setup}

\noindent\textbf{Database.}
The study used the MIMIC-IV Clinical Database Demo dataset~\cite{PhysioNet-mimic-iv-demo-2.2,goldberger2000physiobank}, a publicly available, de-identified subset of MIMIC-IV~\cite{johnson2023mimic} containing structured clinical data for 100 Intensive Care Unit (ICU) patients. 
The dataset includes structured EHR data such as admissions, ICU stays, and procedures, but does not include free-text clinical notes.

\noindent\textbf{Agent Model.}
During the main study, GPT-5.4~\cite{gpt5.4} was selected as the text-to-SQL agent for both interfaces based on its low response latency and stable tool-calling performance.
The agent interpreted natural-language queries, generated and executed SQL against the database, and processed the retrieved results for presentation to the user.
To avoid model-specific expectation bias, participants were not informed of the underlying model.
More details on the model selection criteria are provided in Appendix~\ref{app:agent_model_evaluation}.

\noindent\textbf{User Interface.}
Participants interacted with a web-based system under identical experimental conditions, differing only in the interface (Appendix~\ref{app:user_study_interface}). 
The single-turn interface followed a conventional one-turn text-to-SQL workflow, converting a user-written natural-language query into SQL and returning an answer in a single interaction.
In contrast, the conversational multi-turn interface allowed users to iteratively refine their natural-language queries through dialogue before SQL generation and could suggest relevant follow-up tasks.
When a query was ambiguous or underspecified, the agent in the single-turn interface independently inferred the user's intent and returned an answer together with an explanation of its interpretation, whereas the agent in the conversational multi-turn interface requested clarification before generating SQL.

\subsection{Pilot Study}
\label{sec:pilot_study}
Before finalizing the main study protocol, we conducted two pilot studies involving a total of seven healthcare professionals.

\subsubsection{First Pilot Study}
\label{sec:pilot_study:first_pilot_study}
The first pilot study involved three healthcare professionals and followed a single experimental procedure.
Participants were given nine example questions, comprising three questions each requiring information from one, two, or three database tables. 
Using a data catalog of the MIMIC-IV Clinical Database Demo dataset schema as a reference, participants then formulated three new questions on-site, one for each query type.
To compare model-level user satisfaction, we evaluated four candidate models (GPT-5 Nano~\cite{gpt5}, GPT-5.2~\cite{gpt5.2}, Gemini 3.0 Pro~\cite{gemini3.0}, and Qwen 3 32B~\cite{yang2025qwen3technicalreport}) under two interfaces (single-turn and conversational multi-turn), resulting in 24 experimental conditions (3 questions $\times$ 4 models $\times$ 2 interfaces).

Several limitations emerged from this pilot study. 
First, because the MIMIC-IV Clinical Database Demo differs from databases commonly used in clinical practice, participants spent considerable time familiarizing themselves with the dataset schema before formulating their own questions. 
Second, evaluating all four models substantially increased the length and repetitiveness of the experiment. 
Differences in response latency across models and repeated testing with the same questions further contributed to participant fatigue and reduced concentration. 
Finally, because participants formulated their own questions rather than selecting from a common question pool, the experimental conditions varied across participants, making it difficult to objectively compare satisfaction and other interface-level feedback under standardized conditions. 
These observations motivated us to adopt a two-phase protocol for the second pilot study.

\subsubsection{Second Pilot Study}
\label{sec:pilot_study:second_pilot_study}
The second pilot study involved four healthcare professionals and was conducted to address the limitations identified in the first pilot study. 
Based on the findings, the experimental procedure was redesigned into two phases.

The first phase was designed to familiarize participants with both the MIMIC-IV Clinical Database Demo dataset and the two interfaces while providing a controlled environment for interface comparison. 
To reduce the time and variability associated with participant-generated questions, we prepared 45 predefined questions with corresponding ground-truth answers in advance. 
To further reduce experiment duration and participant fatigue, we evaluated only GPT-5.2, which demonstrated the best latency and stability in the first pilot study; participants were blinded to the model identity. 
To ensure comparable conditions across interfaces, subsequent interactions were disabled in the conversational multi-turn interface during Phase 1. 
Participants evaluated only the initial response from each interface, including any clarification requests or suggested follow-up analyses generated by the conversational interface, and provided descriptive feedback.
The second phase was designed to evaluate interface use under more realistic conditions after participants had become familiar with the system.
Participants were allowed to freely explore the database with queries reflecting information needs arising from their clinical practice or research, without restrictions on query format or source.
After completing the study, participants completed the post-study evaluations.

Feedback from the second pilot study indicated that restricting follow-up interaction in the conversational interface during Phase 1 prevented participants from fully experiencing its interaction characteristics. 
Accordingly, follow-up interactions were enabled during Phase 1 of the main study, while the two-phase protocol was retained.

\subsection{Main Study Procedures}
\label{sec:main_procedures}
An overview of the main study process, finalized following two pilot studies, is shown in Fig.~\ref{fig:design}.

\noindent\textbf{Background Questionnaire.}
Before the experiment, participants completed a background questionnaire assessing their frequency of EHR (or EMR)/CDW use, proficiency in SQL and statistical analysis tools, and use of generative AI in their clinical or daily work.

\noindent\textbf{Phase 1 (Controlled Comparison).}
Phase 1 served as both a familiarization stage and a controlled comparison, with participants selecting queries from a predefined set while experiencing the distinct interaction modes of the two interfaces.
Participants were presented with 45 predefined clinical queries (FX-Q, Appendix~\ref{app:survey_details}) designed for the MIMIC-IV Demo dataset, comprising three sets of 15 queries that required consulting one, two, or three tables, respectively, to answer.
Participants selected one query from each category (three in total) and used each selected query with both interfaces.
After completing Phase 1, they provided query-level open-ended feedback on their satisfaction with the responses from each interface.

\noindent\textbf{Phase 2 (Ecological Validity).}
In Phase 2, participants explored the data in a more realistic setting without restrictions on query format.
They could freely formulate questions reflecting their own clinical or research information needs.
This phase complemented the controlled comparison in Phase 1 by examining use of the two interfaces during open-ended data exploration and enabled analysis of naturally formulated queries from participants.

\noindent\textbf{Post-study Evaluation.}
After completing both phases, participants evaluated each interface using the SUS and UTAUT questionnaires.
They also completed the QDL assessment across three clinical information retrieval scenarios, with higher scores indicating more detailed query formulation.
Finally, participants completed three forced-choice questions comparing the interfaces in terms of overall satisfaction, willingness to adopt, and perceived trustworthiness, and provided open-ended explanations for their choices.
The complete questionnaires are provided in Appendix~\ref{app:survey_details}.

\subsection{Usability Analysis}
\subsubsection{Quantitative Score Computation}
\label{sec:method:usability_analysis:quantitative_score_computation}
\noindent\textbf{SUS.}
SUS responses were scored according to the standardized procedure by Brooke~\cite{brooke1996sus}, yielding a composite score ranging from 0 to 100 for each participant and interface condition (Appendix~\ref{app:sus}).

\noindent\textbf{UTAUT.}
The 19 UTAUT items were grouped into five constructs: Performance Expectancy (PE), Effort Expectancy (EE), Self-Efficacy (SE), Anxiety (ANX), and Behavioral Intention (BI), with each construct score calculated as the mean of its constituent items (Appendix~\ref{app:utaut}).
Because higher ANX scores indicate greater apprehension, ANX was analyzed separately.
We additionally calculated Avg.\ UTAUT (P) as the mean of the four positively oriented constructs (PE, EE, SE, and BI).

\subsubsection{Statistical Comparison Procedure}
\label{sec:method:usability_analysis:statistical_comparison_procedure}
Given the within-subjects design, usability scores from the two interfaces were analyzed as paired observations.
For each metric, normality of the paired differences was assessed using the Shapiro--Wilk test~\cite{shaphiro1965analysis}.
A paired $t$-test was used as the primary inferential test when the normality assumption was satisfied; otherwise, the Wilcoxon signed-rank test~\cite{wilcoxon1945individual} was used.
We report descriptive statistics, $p$-values, and corresponding effect sizes (Cohen's $d_z$~\cite{cohen2013statistical} for paired $t$-tests and $r$ for Wilcoxon tests), with additional statistical details provided in Appendix~\ref{app:statistical_analysis_details}.

\section*{Data availability}
The MIMIC-IV Clinical Database Demo (v2.2) used in this study is publicly available on PhysioNet (\url{https://physionet.org/content/mimic-iv-demo/2.2/}). 
De-identified dialogue logs and survey responses collected during the user study are planned for public release. 
Until then, they are available from the corresponding author upon reasonable request, subject to approval by the Institutional Review Board of Samsung Medical Center.

\section*{Code availability}
The code for the text-to-SQL interfaces and analysis scripts is available from the corresponding author upon reasonable request.

\section*{Acknowledgements}
This research was supported by the Institute of Information \& Communications Technology Planning \& Evaluation (IITP) grants (No.RS-2019-II190075, No.RS-2025-02304967), the Korea Health Industry Development Institute (KHIDI) grant (No.RS-2025-02223382) and the National Research Foundation of Korea (NRF) grants (NRF-2020H1D3A2A03100945, RS-2026-25484088), funded by the Korea government (MSIT, MOHW).
We thank Yoon Jae Kim (Department of Electrical and Computer Engineering, Sungkyunkwan University) for contributing to the experimental design, conducting the user study experiments, and constructing the ground-truth dataset.

\section*{Author information}
\subsection*{Author Affiliations}
\noindent\textbf{Department of Kim Jaechul Graduate School of AI, Korea Advanced Institute of Science and Technology, Seongnam-si, 13558, Gyeonggi-do, Republic of Korea}

\noindent Jun-Min Lee, Gyubok Lee \& Edward Choi
\vspace{1em}

\noindent\textbf{Research Institute for Future Medicine, Samsung Medical Center, Seoul, 06351, Republic of Korea}

\noindent Suyoung Yoo
\vspace{1em}

\noindent\textbf{Department of Emergency Medicine, Samsung Medical Center, Seoul, 06351, Republic of Korea}

\noindent Meong Hi Son
\vspace{1em}

\noindent\textbf{Office of AI Transformation, Samsung Medical Center, Seoul, 06351, Republic of Korea}

\noindent Meong Hi Son
\vspace{1em}

\subsection*{Author Contributions}
J.-M.L. led the overall conceptualization of the study, developed and implemented the machine learning pipeline, conducted the user study experiments and formal analyses, prepared the tables and figures, and drafted the manuscript.
S.Y. contributed to the study design and coordination, supported data preparation and clinical-domain interpretation, prepared the figures, and reviewed and revised the manuscript.
G.L. contributed to the implementation of the machine learning pipeline and experimental design.
E.C. supervised the study, provided guidance on the development of the machine learning pipeline and experimental design, and reviewed and revised the manuscript.
M.H.S. supervised the study, provided clinical guidance and resources, contributed to the conceptualization and interpretation of the results, and reviewed and revised the manuscript.
All authors reviewed and approved the final version of the manuscript.

\section*{Ethics Declarations}
The study protocol was approved by the Institutional Review Board of Samsung Medical Center (approval no.~2026-02-010). All participants provided written informed consent.

\subsection*{Competing Interests}
The authors declare no competing interests.


\bibliography{sn-bibliography}

\begin{thebibliography}{10}
\expandafter\ifx\csname url\endcsname\relax
  \def\url#1{\burl{#1}}\fi
\expandafter\ifx\csname urlprefix\endcsname\relax\def\urlprefix{URL }\fi
\providecommand{\bibinfo}[2]{#2}
\providecommand{\eprint}[2][]{\url{#2}}
\providecommand{\doi}[1]{\url{https://doi.org/#1}}
\bibcommenthead

\bibitem{shaphiro1965analysis-app}
\bibinfo{author}{Shaphiro, S.}, \bibinfo{author}{Wilk, M.} \emph{et~al.}
\newblock \bibinfo{title}{An analysis of variance test for normality}.
\newblock \emph{\bibinfo{journal}{Biometrika}} \textbf{\bibinfo{volume}{52}}, \bibinfo{pages}{591--611} (\bibinfo{year}{1965}).

\bibitem{wilcoxon1945individual-app}
\bibinfo{author}{Wilcoxon, F.}
\newblock \bibinfo{title}{Individual comparisons by ranking methods}.
\newblock \emph{\bibinfo{journal}{Biometrics bulletin}} \textbf{\bibinfo{volume}{1}}, \bibinfo{pages}{80--83} (\bibinfo{year}{1945}).

\bibitem{gpt5-app}
\bibinfo{author}{{OpenAI}}.
\newblock \bibinfo{title}{Introducing gpt-5} (\bibinfo{year}{2025}).
\newblock \urlprefix\url{https://openai.com/index/introducing-gpt-5/}.
\newblock \bibinfo{note}{Accessed: 2026-03-05}.

\bibitem{gpt5.2-app}
\bibinfo{author}{{OpenAI}}.
\newblock \bibinfo{title}{Introducing gpt-5.2} (\bibinfo{year}{2025}).
\newblock \urlprefix\url{https://openai.com/index/introducing-gpt-5-2/}.
\newblock \bibinfo{note}{Accessed: 2026-03-05}.

\bibitem{gemini3.0-app}
\bibinfo{author}{{Google Cloud}}.
\newblock \bibinfo{title}{Gemini 3 pro} (\bibinfo{year}{2026}).
\newblock \urlprefix\url{https://docs.cloud.google.com/vertex-ai/generative-ai/docs/models/gemini/3-pro}.
\newblock \bibinfo{note}{Accessed: 2026-03-05}.

\bibitem{yang2025qwen3technicalreport-app}
\bibinfo{author}{Yang, A.} \emph{et~al.}
\newblock \bibinfo{title}{Qwen3 technical report} (\bibinfo{year}{2025}).
\newblock \urlprefix\url{https://arxiv.org/abs/2505.09388}.
\newblock \bibinfo{eprint}{{\href{https://arxiv.org/abs/2505.09388}{{arXiv:2505.09388}}}}.

\bibitem{gpt5.4-app}
\bibinfo{author}{{OpenAI}}.
\newblock \bibinfo{title}{Introducing gpt-5.4} (\bibinfo{year}{2026}).
\newblock \urlprefix\url{https://openai.com/index/introducing-gpt-5-4/}.
\newblock \bibinfo{note}{Accessed: 2026-03-05}.

\bibitem{brooke1996sus-app}
\bibinfo{author}{Brooke, J.} \emph{et~al.}
\newblock \bibinfo{title}{Sus-a quick and dirty usability scale}.
\newblock \emph{\bibinfo{journal}{Usability evaluation in industry}} \textbf{\bibinfo{volume}{189}}, \bibinfo{pages}{4--7} (\bibinfo{year}{1996}).

\bibitem{cohen2013statistical-app}
\bibinfo{author}{Cohen, J.}
\newblock \emph{\bibinfo{title}{Statistical power analysis for the behavioral sciences}}  (\bibinfo{publisher}{routledge}, \bibinfo{year}{2013}).

\bibitem{brydges2019effect-app}
\bibinfo{author}{Brydges, C.~R.}
\newblock \bibinfo{title}{Effect size guidelines, sample size calculations, and statistical power in gerontology}.
\newblock \emph{\bibinfo{journal}{Innovation in aging}} \textbf{\bibinfo{volume}{3}}, \bibinfo{pages}{igz036} (\bibinfo{year}{2019}).

\end{thebibliography}


\begin{thebibliography}{10}
\expandafter\ifx\csname url\endcsname\relax
  \def\url#1{\burl{#1}}\fi
\expandafter\ifx\csname urlprefix\endcsname\relax\def\urlprefix{URL }\fi
\providecommand{\bibinfo}[2]{#2}
\providecommand{\eprint}[2][]{\url{#2}}
\providecommand{\doi}[1]{\url{https://doi.org/#1}}
\bibcommenthead

\bibitem{charles2015adoption}
\bibinfo{author}{Charles, D.}, \bibinfo{author}{Gabriel, M.} \& \bibinfo{author}{Searcy, T.}
\newblock \bibinfo{title}{Adoption of electronic health record systems among us non-federal acute care hospitals: 2008--2014}.
\newblock \emph{\bibinfo{journal}{ASTP Health IT Data Brief [Internet]}}  (\bibinfo{year}{2015}).

\bibitem{jensen2012mining}
\bibinfo{author}{Jensen, P.~B.}, \bibinfo{author}{Jensen, L.~J.} \& \bibinfo{author}{Brunak, S.}
\newblock \bibinfo{title}{Mining electronic health records: towards better research applications and clinical care}.
\newblock \emph{\bibinfo{journal}{Nature Reviews Genetics}} \textbf{\bibinfo{volume}{13}}, \bibinfo{pages}{395--405} (\bibinfo{year}{2012}).

\bibitem{holmes2021electronic}
\bibinfo{author}{Holmes, J.~H.} \emph{et~al.}
\newblock \bibinfo{title}{Why is the electronic health record so challenging for research and clinical care?}
\newblock \emph{\bibinfo{journal}{Methods of information in medicine}} \textbf{\bibinfo{volume}{60}}, \bibinfo{pages}{032--048} (\bibinfo{year}{2021}).

\bibitem{miller2024automated}
\bibinfo{author}{Miller, T.~P.} \emph{et~al.}
\newblock \bibinfo{title}{Automated electronic health record data extraction and curation using extractehr}.
\newblock \emph{\bibinfo{journal}{JCO Clinical Cancer Informatics}} \textbf{\bibinfo{volume}{8}}, \bibinfo{pages}{e2400100} (\bibinfo{year}{2024}).

\bibitem{olakotan2025usability}
\bibinfo{author}{Olakotan, O.}, \bibinfo{author}{Samuriwo, R.}, \bibinfo{author}{Ismaila, H.} \& \bibinfo{author}{Atiku, S.}
\newblock \bibinfo{title}{Usability challenges in electronic health records: impact on documentation burden and clinical workflow: a scoping review}.
\newblock \emph{\bibinfo{journal}{Journal of Evaluation in Clinical Practice}} \textbf{\bibinfo{volume}{31}}, \bibinfo{pages}{e70189} (\bibinfo{year}{2025}).

\bibitem{zhong2017seq2sql}
\bibinfo{author}{Zhong, V.}, \bibinfo{author}{Xiong, C.} \& \bibinfo{author}{Socher, R.}
\newblock \bibinfo{title}{Seq2sql: Generating structured queries from natural language using reinforcement learning} (\bibinfo{year}{2017}).
\newblock \urlprefix\url{https://arxiv.org/abs/1709.00103}.
\newblock \bibinfo{eprint}{{\href{https://arxiv.org/abs/1709.00103}{{arXiv:1709.00103}}}}.

\bibitem{yu2018spider}
\bibinfo{author}{Yu, T.} \emph{et~al.} \emph{\bibinfo{title}{Spider: A large-scale human-labeled dataset for complex and cross-domain semantic parsing and text-to-sql task}}.
\newblock \emph{\bibinfo{booktitle}{Proceedings of the 2018 Conference on Empirical Methods in Natural Language Processing (EMNLP)}}, \bibinfo{pages}{3911--3921} (\bibinfo{year}{2018}).

\bibitem{pourreza2023dinsql}
\bibinfo{author}{Pourreza, M.} \& \bibinfo{author}{Rafiei, D.}
\newblock \bibinfo{title}{Din-sql: Decomposed in-context learning of text-to-sql with self-correction}.
\newblock \emph{\bibinfo{journal}{Advances in Neural Information Processing Systems}} \textbf{\bibinfo{volume}{36}}, \bibinfo{pages}{36339--36348} (\bibinfo{year}{2023}).

\bibitem{hong2024nextgeneration}
\bibinfo{author}{Hong, Z.} \emph{et~al.}
\newblock \bibinfo{title}{Next-generation database interfaces: A survey of llm-based text-to-sql} (\bibinfo{year}{2024}).
\newblock \urlprefix\url{https://arxiv.org/abs/2406.08426}.
\newblock \bibinfo{eprint}{{\href{https://arxiv.org/abs/2406.08426}{{arXiv:2406.08426}}}}.

\bibitem{wang2020text}
\bibinfo{author}{Wang, P.}, \bibinfo{author}{Shi, T.} \& \bibinfo{author}{Reddy, C.~K.} \emph{\bibinfo{title}{Text-to-sql generation for question answering on electronic medical records}}.
\newblock \emph{\bibinfo{booktitle}{Proceedings of The Web Conference 2020}}, \bibinfo{pages}{350--361} (\bibinfo{year}{2020}).

\bibitem{lee2022ehrsql}
\bibinfo{author}{Lee, G.} \emph{et~al.}
\newblock \bibinfo{title}{Ehrsql: A practical text-to-sql benchmark for electronic health records}.
\newblock \emph{\bibinfo{journal}{Advances in Neural Information Processing Systems}} \textbf{\bibinfo{volume}{35}}, \bibinfo{pages}{15589--15601} (\bibinfo{year}{2022}).

\bibitem{ryu2024ehr}
\bibinfo{author}{Ryu, J.}, \bibinfo{author}{Cho, S.}, \bibinfo{author}{Lee, G.} \& \bibinfo{author}{Choi, E.} \emph{\bibinfo{title}{Ehr-seqsql: A sequential text-to-sql dataset for interactively exploring electronic health records}}.
\newblock \emph{\bibinfo{booktitle}{Findings of the Association for Computational Linguistics: {ACL} 2024}}, \bibinfo{pages}{16388--16407} (\bibinfo{year}{2024}).

\bibitem{ohno2011realizing}
\bibinfo{author}{Ohno-Machado, L.}
\newblock \bibinfo{title}{Realizing the full potential of electronic health records: the role of natural language processing}.
\newblock \emph{\bibinfo{journal}{Journal of the American Medical Informatics Association}} \textbf{\bibinfo{volume}{18}}, \bibinfo{pages}{539--539} (\bibinfo{year}{2011}).

\bibitem{yang2022large}
\bibinfo{author}{Yang, X.} \emph{et~al.}
\newblock \bibinfo{title}{A large language model for electronic health records}.
\newblock \emph{\bibinfo{journal}{NPJ digital medicine}} \textbf{\bibinfo{volume}{5}}, \bibinfo{pages}{194} (\bibinfo{year}{2022}).

\bibitem{3666122.3667957}
\bibinfo{author}{Li, J.} \emph{et~al.} \emph{\bibinfo{title}{Can llm already serve as a database interface? a big bench for large-scale database grounded text-to-sqls}}.
\newblock \emph{\bibinfo{booktitle}{Proceedings of the 37th International Conference on Neural Information Processing Systems}}, NIPS '23 (\bibinfo{publisher}{Curran Associates Inc.}, \bibinfo{address}{Red Hook, NY, USA}, \bibinfo{year}{2023}).

\bibitem{ipeirotis2025nlidb}
\bibinfo{author}{Ipeirotis, P.} \& \bibinfo{author}{Zheng, H.}
\newblock \bibinfo{title}{Natural language interfaces for databases: What do users think?} (\bibinfo{year}{2025}).
\newblock \urlprefix\url{https://arxiv.org/abs/2511.14718}.
\newblock \bibinfo{eprint}{{\href{https://arxiv.org/abs/2511.14718}{{arXiv:2511.14718}}}}.

\bibitem{min2020ambigqa}
\bibinfo{author}{Min, S.}, \bibinfo{author}{Michael, J.}, \bibinfo{author}{Hajishirzi, H.} \& \bibinfo{author}{Zettlemoyer, L.} \emph{\bibinfo{title}{Ambigqa: Answering ambiguous open-domain questions}}.
\newblock \emph{\bibinfo{booktitle}{Proceedings of the 2020 Conference on Empirical Methods in Natural Language Processing (EMNLP)}}, \bibinfo{pages}{5783--5797} (\bibinfo{year}{2020}).

\bibitem{wang-etal-2023-know}
\bibinfo{author}{Wang, B.}, \bibinfo{author}{Gao, Y.}, \bibinfo{author}{Li, Z.} \& \bibinfo{author}{Lou, J.-G.} \emph{\bibinfo{title}{Know what {I} don{'}t know: Handling ambiguous and unknown questions for text-to-{SQL}}}.
\newblock \emph{\bibinfo{booktitle}{Findings of the Association for Computational Linguistics: ACL 2023}}, \bibinfo{pages}{5701--5714} (\bibinfo{publisher}{Association for Computational Linguistics}, \bibinfo{address}{Toronto, Canada}, \bibinfo{year}{2023}).
\newblock \urlprefix\url{https://aclanthology.org/2023.findings-acl.352/}.

\bibitem{rao2018learning}
\bibinfo{author}{Rao, S.} \& \bibinfo{author}{Daum{\'e}~III, H.} \emph{\bibinfo{title}{Learning to ask good questions: Ranking clarification questions using neural expected value of perfect information}}.
\newblock \emph{\bibinfo{booktitle}{Proceedings of the 56th Annual Meeting of the Association for Computational Linguistics (ACL)}}, \bibinfo{pages}{2737--2746} (\bibinfo{year}{2018}).

\bibitem{yao2019model}
\bibinfo{author}{Yao, Z.}, \bibinfo{author}{Su, Y.}, \bibinfo{author}{Sun, H.} \& \bibinfo{author}{Yih, W.-t.} \emph{\bibinfo{title}{Model-based interactive semantic parsing: A unified framework and a text-to-sql case study}}.
\newblock \emph{\bibinfo{booktitle}{Proceedings of the 2019 Conference on Empirical Methods in Natural Language Processing and the 9th International Joint Conference on Natural Language Processing (EMNLP-IJCNLP)}}, \bibinfo{pages}{5447--5458} (\bibinfo{year}{2019}).

\bibitem{lee2026conversationqueryexecutionbenchmarking}
\bibinfo{author}{Lee, G.} \emph{et~al.}
\newblock \bibinfo{title}{From conversation to query execution: Benchmarking user and tool interactions for ehr database agents} (\bibinfo{year}{2026}).
\newblock \urlprefix\url{https://arxiv.org/abs/2509.23415}.
\newblock \bibinfo{eprint}{{\href{https://arxiv.org/abs/2509.23415}{{arXiv:2509.23415}}}}.

\bibitem{johnson2023mimic}
\bibinfo{author}{Johnson, A.~E.} \emph{et~al.}
\newblock \bibinfo{title}{Mimic-iv, a freely accessible electronic health record dataset}.
\newblock \emph{\bibinfo{journal}{Scientific data}} \textbf{\bibinfo{volume}{10}}, \bibinfo{pages}{1} (\bibinfo{year}{2023}).

\bibitem{brooke1996sus}
\bibinfo{author}{Brooke, J.} \emph{et~al.}
\newblock \bibinfo{title}{Sus-a quick and dirty usability scale}.
\newblock \emph{\bibinfo{journal}{Usability evaluation in industry}} \textbf{\bibinfo{volume}{189}}, \bibinfo{pages}{4--7} (\bibinfo{year}{1996}).

\bibitem{venkatesh2003user}
\bibinfo{author}{Venkatesh, V.}, \bibinfo{author}{Morris, M.~G.}, \bibinfo{author}{Davis, G.~B.} \& \bibinfo{author}{Davis, F.~D.}
\newblock \bibinfo{title}{User acceptance of information technology: Toward a unified view1}.
\newblock \emph{\bibinfo{journal}{MIS quarterly}} \textbf{\bibinfo{volume}{27}}, \bibinfo{pages}{425--478} (\bibinfo{year}{2003}).

\bibitem{bangor2009determining}
\bibinfo{author}{Bangor, A.}, \bibinfo{author}{Kortum, P.} \& \bibinfo{author}{Miller, J.}
\newblock \bibinfo{title}{Determining what individual sus scores mean: Adding an adjective rating scale}.
\newblock \emph{\bibinfo{journal}{Journal of Usability Studies}} \textbf{\bibinfo{volume}{4}}, \bibinfo{pages}{114--123} (\bibinfo{year}{2009}).

\bibitem{brydges2019effect}
\bibinfo{author}{Brydges, C.~R.}
\newblock \bibinfo{title}{Effect size guidelines, sample size calculations, and statistical power in gerontology}.
\newblock \emph{\bibinfo{journal}{Innovation in aging}} \textbf{\bibinfo{volume}{3}}, \bibinfo{pages}{igz036} (\bibinfo{year}{2019}).

\bibitem{PhysioNet-mimic-iv-demo-2.2}
\bibinfo{author}{Johnson, A.} \emph{et~al.}
\newblock \bibinfo{title}{{MIMIC-IV Clinical Database Demo}}.
\newblock \emph{\bibinfo{journal}{{PhysioNet}}}  (\bibinfo{year}{2023}).
\newblock \urlprefix\url{https://doi.org/10.13026/dp1f-ex47}.
\newblock \bibinfo{note}{Version 2.2}.

\bibitem{goldberger2000physiobank}
\bibinfo{author}{Goldberger, A.~L.} \emph{et~al.}
\newblock \bibinfo{title}{Physiobank, physiotoolkit, and physionet: components of a new research resource for complex physiologic signals}.
\newblock \emph{\bibinfo{journal}{circulation}} \textbf{\bibinfo{volume}{101}}, \bibinfo{pages}{e215--e220} (\bibinfo{year}{2000}).

\bibitem{gpt5.4}
\bibinfo{author}{{OpenAI}}.
\newblock \bibinfo{title}{Introducing gpt-5.4} (\bibinfo{year}{2026}).
\newblock \urlprefix\url{https://openai.com/index/introducing-gpt-5-4/}.
\newblock \bibinfo{note}{Accessed: 2026-03-05}.

\bibitem{gpt5}
\bibinfo{author}{{OpenAI}}.
\newblock \bibinfo{title}{Introducing gpt-5} (\bibinfo{year}{2025}).
\newblock \urlprefix\url{https://openai.com/index/introducing-gpt-5/}.
\newblock \bibinfo{note}{Accessed: 2026-03-05}.

\bibitem{gpt5.2}
\bibinfo{author}{{OpenAI}}.
\newblock \bibinfo{title}{Introducing gpt-5.2} (\bibinfo{year}{2025}).
\newblock \urlprefix\url{https://openai.com/index/introducing-gpt-5-2/}.
\newblock \bibinfo{note}{Accessed: 2026-03-05}.

\bibitem{gemini3.0}
\bibinfo{author}{{Google Cloud}}.
\newblock \bibinfo{title}{Gemini 3 pro} (\bibinfo{year}{2026}).
\newblock \urlprefix\url{https://docs.cloud.google.com/vertex-ai/generative-ai/docs/models/gemini/3-pro}.
\newblock \bibinfo{note}{Accessed: 2026-03-05}.

\bibitem{yang2025qwen3technicalreport}
\bibinfo{author}{Yang, A.} \emph{et~al.}
\newblock \bibinfo{title}{Qwen3 technical report} (\bibinfo{year}{2025}).
\newblock \urlprefix\url{https://arxiv.org/abs/2505.09388}.
\newblock \bibinfo{eprint}{{\href{https://arxiv.org/abs/2505.09388}{{arXiv:2505.09388}}}}.

\bibitem{shaphiro1965analysis}
\bibinfo{author}{Shaphiro, S.}, \bibinfo{author}{Wilk, M.} \emph{et~al.}
\newblock \bibinfo{title}{An analysis of variance test for normality}.
\newblock \emph{\bibinfo{journal}{Biometrika}} \textbf{\bibinfo{volume}{52}}, \bibinfo{pages}{591--611} (\bibinfo{year}{1965}).

\bibitem{wilcoxon1945individual}
\bibinfo{author}{Wilcoxon, F.}
\newblock \bibinfo{title}{Individual comparisons by ranking methods}.
\newblock \emph{\bibinfo{journal}{Biometrics bulletin}} \textbf{\bibinfo{volume}{1}}, \bibinfo{pages}{80--83} (\bibinfo{year}{1945}).

\bibitem{cohen2013statistical}
\bibinfo{author}{Cohen, J.}
\newblock \emph{\bibinfo{title}{Statistical power analysis for the behavioral sciences}}  (\bibinfo{publisher}{routledge}, \bibinfo{year}{2013}).

\end{thebibliography}

\clearpage
\setcounter{page}{1}
\begin{appendices}

\renewcommand{\thetable}{S\arabic{table}}
\renewcommand{\thefigure}{S\arabic{figure}}

\section{Supplementary Results and Analysis}
\label{app:supplementary_results}

\subsection{Supplementary Statistical Results}
\label{app:supplementary_results:supplementary_statistical_results}
\begin{table*}[ht!]
    \caption{Paired comparison of SUS and UTAUT outcomes between the single-turn and conversational multi-turn interfaces ($n=14$). SD denotes standard deviation. $p_{\mathrm{SW}}$ denotes the Shapiro--Wilk test $p$-value for the paired differences. $d_z$ and $r$ denote Cohen's $d_z$ and the Wilcoxon effect size, respectively. n.s., not significant.}
    \label{tab:statistical_results}
        \resizebox{1\linewidth}{!}{
            \begin{tabular}{lccc|ccc|ccc}
            \toprule
            \multirow[c]{2}{*}{\textbf{Metric}} & \multirow[c]{2}{*}{\makecell[c]{\textbf{Single-turn}\\\textbf{Mean (SD)}}} & \multirow[c]{2}{*}{\makecell[c]{\textbf{Multi-turn}\\\textbf{Mean (SD)}}} & \multirow[c]{2}{*}{\makecell[c]{\textbf{Shapiro--Wilk}\\$\boldsymbol{p_{\mathrm{SW}}}$}} & \multicolumn{3}{c|}{\textbf{Paired $t$-test}} & \multicolumn{3}{c}{\textbf{Wilcoxon}} \\
            \cmidrule(lr){5-7}\cmidrule(lr){8-10}
            & & & & \textbf{$t$(13)} & \textbf{$p$} & \textbf{$d_z$} & \textbf{$Z$} & \textbf{$p$} & \textbf{$r$} \\
            \midrule
            SUS              & 68.2 (16.7) & 85.9 (11.8) & 0.176 & $-5.09$ & $<0.001$     & $-1.36$ & $-3.30$ & 0.001        & $-0.88$ \\
            PE               & 3.16 (1.24) & 4.82 (0.32) & \textbf{0.045}$^{\dagger}$ & $-5.03$ & $<0.001$     & $-1.34$  & $-3.30$   & 0.001        & $-0.88$ \\
            EE               & 4.04 (0.87) & 4.52 (0.44) & \textbf{0.017}$^{\dagger}$ & $-2.82$ & 0.014        & $-0.75$ & $-2.55$ & 0.010        & $-0.85$ \\
            SE               & 3.91 (0.96) & 4.55 (0.38) & \textbf{0.002}$^{\dagger}$ & $-2.68$ & 0.019        & $-0.72$ & $-2.37$  & 0.018        & $-0.89$ \\
            ANX              & 1.89 (0.94) & 1.70 (0.65) & 0.052 & $+1.36$ & 0.196 (n.s.) & -     & $+1.27$  & 0.201 (n.s.) & - \\
            BI               & 2.67 (1.28) & 4.62 (0.57) & 0.224 & $-5.13$ & $<0.001$     & $-1.37$ & $-3.06$  & 0.002        & $-0.88$ \\
            Avg.\ UTAUT (P)  & 3.44 (0.84) & 4.63 (0.34) & 0.177 & $-5.16$ & $<0.001$     & $-1.38$ & $-3.30$  & $<0.001$     & $-0.88$ \\
            \bottomrule
            \end{tabular}
        }

    \vspace{0.4em}
    \begin{minipage}{0.95\linewidth}
    \footnotesize
    \textit{Note:} $^{\dagger}$Normality assumption violated ($p_{\mathrm{SW}}<0.05$); the Wilcoxon signed-rank test was treated as the primary test for this metric (Section~\ref{sec:method:usability_analysis:statistical_comparison_procedure}).
    \end{minipage}
\end{table*}

\noindent Table~\ref{tab:statistical_results} provides the complete descriptive and inferential statistics underlying the usability comparison.
For each metric, the table reports the mean and standard deviation for each interface, the Shapiro--Wilk test~\citeapp{shaphiro1965analysis-app} for assessing the normality of the paired differences, and the results of both the paired $t$-test and the Wilcoxon signed-rank test~\citeapp{wilcoxon1945individual-app}, together with their corresponding effect sizes.

Overall, the two inferential tests yielded consistent results across all metrics.
As described in Section~\ref{sec:method:usability_analysis:statistical_comparison_procedure}, the Wilcoxon signed-rank test was treated as the primary inferential test when the normality assumption was violated ($p_{\mathrm{SW}}<0.05$).
Consistent with the main findings, statistically significant differences favoring the conversational multi-turn interface were observed for all usability and technology-acceptance measures except Anxiety (ANX), with most significant comparisons showing large effect sizes.
ANX did not differ significantly between the two interfaces, and both yielded low absolute scores, with the conversational multi-turn interface showing a slightly lower mean (lower scores indicate lower anxiety).

\subsection{Query Detail Level Distribution}
\label{app:qdl_distribution}

\begin{figure}[t!]
    \centering
    \includegraphics[width=\textwidth]{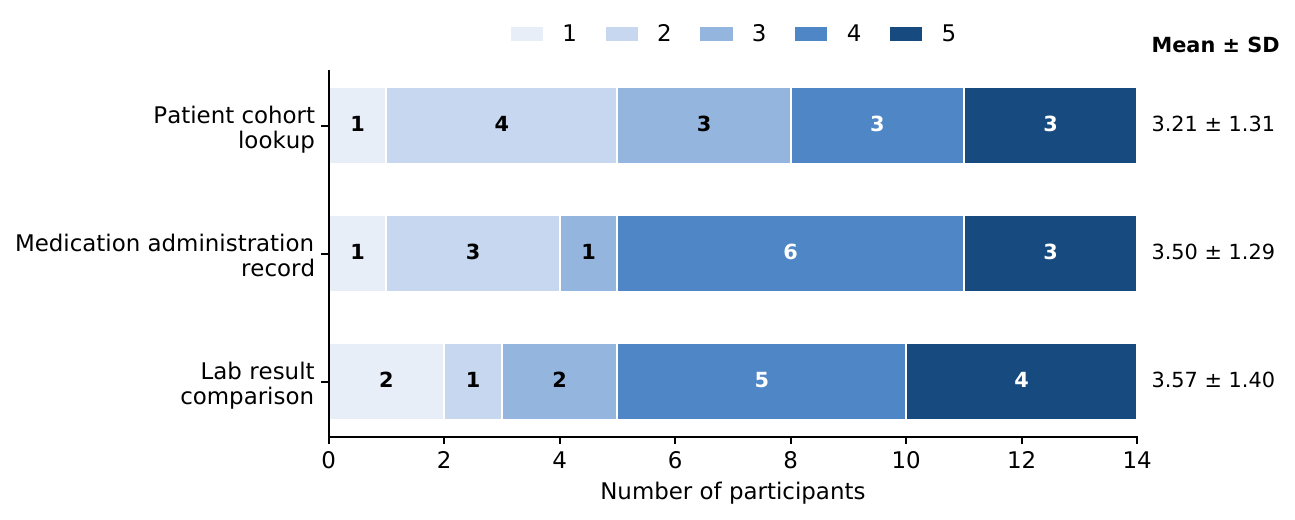}
        \caption{Distribution of self-reported Query Detail Level (QDL) ratings across the three clinical information-seeking scenarios ($n=14$). Ratings ranged from 1 (vague example phrasing) to 5 (detailed example phrasing). Detailed scenarios and example phrasings are provided in Table~\ref{tab:survey}.}
    \label{fig:qdl_distribution}
\end{figure}

Query Detail Level (QDL) was assessed across three clinical information-seeking scenarios in which participants rated how specifically they would formulate a natural-language query on a five-point scale.
As shown in Fig.~\ref{fig:qdl_distribution}, ratings varied across both participants and scenarios, indicating substantial individual variation in query formulation style.
Participant-level mean QDL scores, calculated by averaging ratings across the three scenarios, ranged from 1.0 to 5.0 ($\mu=3.43$, $\sigma=1.07$).
This broad distribution indicates that participants differed considerably in their tendency to formulate detailed versus underspecified queries.

\subsection{Dialogue Turn Definitions and Distribution}
\label{app:supplementary_results:turn_definition}
\begin{table*}[t!]
    \caption{Definitions of the seven turn types used to annotate dialogue turns.}
    \label{tab:turn_type_definitions}
    \centering
    \resizebox{1\linewidth}{!}{
    \begin{tabular}{l l p{9cm}}
    \toprule
    \textbf{Origin} & \textbf{Type} & \textbf{Definition} \\
    \midrule
    \multirow{4}{*}{User}
        & Question (Q) & A query freely composed by the participant, without predefined constraints on its format or scope. \\
        & Fixed Question (FX-Q) & A predefined, highly specified question with an expected answer established in advance by the study investigators. \\
        & Suggested Follow-up Question (SF-Q) & A user turn selecting one of the follow-up analysis options suggested by the agent after presenting a result. \\
        & Clarification Response (CL-Q) & A user turn providing additional information in response to an agent clarification request, thereby refining the preceding query. \\
    \midrule
    \multirow{3}{*}{Agent}
        & Answer (ANS) & An agent response containing neither a clarification request nor a follow-up suggestion. \\
        & Answer with Suggestion (S-ANS) & An agent response presenting a result together with one or more suggested follow-up analyses. If a suggestion is selected, the subsequent user turn is labeled SF-Q. \\
        & Clarification Request (C-ANS) & An agent response requesting additional information or clarification before providing an answer. If the user responds, the subsequent user turn is labeled CL-Q. \\
    \bottomrule
    \end{tabular}
    }
\end{table*}
\begin{table*}[t!]
    \caption{Distribution of annotated user-turn and agent-response types by interface.}
    \label{tab:turn_counts}
    \footnotesize
    \centering
    \begin{tabular}{lrr}
    \toprule

    \multicolumn{3}{l}{\textbf{User Turn}} \\
    \midrule
    \textbf{Turn Type} 
        & \textbf{Single-turn} 
        & \textbf{Conversational Multi-turn} \\
    \midrule
    Question (Q)                     & 40 (39.2\%)   & 161 (49.8\%) \\
    Fixed Question (FX-Q)            & 62 (60.8\%)   & 56 (17.3\%) \\
    Suggested Follow-up Question (SF-Q) & --          & 60 (18.6\%) \\
    Clarification Response (CL-Q)    & --            & 46 (14.2\%) \\
    \addlinespace
    \textbf{Total}                   & \textbf{102 (100.0\%)} & \textbf{323 (100.0\%)} \\

    \midrule

    \multicolumn{3}{l}{\textbf{Agent Response}} \\
    \midrule
    \textbf{Response Type} 
        & \textbf{Single-turn} 
        & \textbf{Conversational Multi-turn} \\
    \midrule
    Answer (ANS)                     & 102 (100.0\%) & 25 (7.8\%) \\
    Answer with Suggestion (S-ANS)   & --            & 249 (77.3\%) \\
    Clarification Request (C-ANS)    & --            & 48 (14.9\%) \\
    \addlinespace
    \textbf{Total}                   & \textbf{102 (100.0\%)} & \textbf{322 (100.0\%)} \\

    \bottomrule
    \end{tabular}

    \vspace{0.4em}
    \begin{minipage}{0.95\linewidth}
    \footnotesize
    \textit{Note:} Percentages are calculated within each interface condition. Dashes indicate turn types not supported by the single-turn interface. One accepted SF-Q has no corresponding agent response because the page was refreshed before the response was returned; therefore, the conversational multi-turn interface contains 323 user turns and 322 agent responses.
    \end{minipage}
\end{table*}

For the turn-level dialogue analysis, each user turn and agent response was categorized according to its conversational function.
The taxonomy comprised four user-turn types and three agent-response types, covering initial queries, clarification exchanges, and follow-up interactions.
Table~\ref{tab:turn_type_definitions} summarizes the definition of each turn type.
Table~\ref{tab:turn_counts} shows the distribution of user and agent-response turn types across the single-turn and conversational multi-turn interfaces.
Because the single-turn interface supports only one-turn interactions, clarification (CL-Q/C-ANS) and follow-up suggestion (SF-Q/S-ANS) turn types occurred exclusively in the conversational multi-turn interface.

\subsection{Clarification Trigger Typology and Coding Criteria}
\label{app:supplementary_results:clarification_typology}
\begin{table*}[t!]
    \caption{Coding criteria for the six qualitative categories of clarification triggers.  Categories shown in italics were considered related to EHR- or clinical-domain knowledge.}
    \label{tab:typology_criteria}
    \centering
    \resizebox{1\linewidth}{!}{
    \begin{tabular}{l p{10cm}}
    \toprule
    \textbf{Category} & \textbf{Coding criterion} \\
    \midrule
    \textit{Cohort/population definition} &
    The criteria defining the target patient population (\eg admission timing, inclusion/exclusion criteria, diagnostic scope) can be interpreted in multiple ways (\texttt{WHERE}-clause, row-wise ambiguity). \\
    
    \textit{Operational definition of a measure} &
    An everyday-language expression (\eg ``regional distribution,'' ``cause of death,'' ``increasing trend'') requires a concrete computational rule to translate into a SQL expression (\texttt{SELECT}-clause, column-wise ambiguity). \\
    
    Aggregation unit/grouping criterion &
    The unit of analysis (\eg patient vs.\ visit) or grouping criterion is unclear. \\
    
    Feasibility of the requested DB/analytic method &
    The requested analysis cannot be performed using the available database, or the proposed statistical method requires revision. \\
    
    \textit{Measurement source/data provenance} &
    The same clinical concept can be derived from multiple data sources or database tables. \\
    
    Anaphora/contextual reference &
    The query refers to an entity mentioned earlier in the conversation and cannot be interpreted without dialogue history. \\
    \bottomrule
    \end{tabular}
    }
\end{table*}
\begin{table*}[t!]
\caption{Qualitative classification of the 42 clarification exchanges by the dominant trigger of the initial clarification request, with representative user queries and ambiguity details. Categories shown in italics were considered related to EHR- or clinical-domain knowledge.}
\label{tab:typology_results}
\centering
\resizebox{\linewidth}{!}{
\begin{tabular}{l r p{5.5cm} p{6cm}}
\toprule
\textbf{Category} & \textbf{Count (\%)} & \textbf{Representative User Query} & \textbf{Source of Ambiguity} \\
\midrule
    \textit{Cohort/population definition}
    & 15 (35.7\%)
    & \textit{``Patients most recently admitted to the ICU for pneumonia\ldots''}
    & What defines ``most recently,'' and whether the lookback window should be anchored to the admission date or the current date. \\

    \textit{Operational definition of a measure}
    & 10 (23.8\%)
    & \textit{``I want to know the regional distribution of patients who visited the ED.''}
    & What ``regional distribution'' refers to, such as residence, race, referral pathway, or insurance, and whether counts should be patient- or visit-based. \\

    Aggregation unit/grouping criterion
    & 8 (19.0\%)
    & \textit{``Recalculate this for in-hospital deaths, and organize the distribution of the time interval between admission and death by age group.''}
    & The ambiguity lies in the aggregation level: whether to report summary statistics by age group, a distribution across time intervals, or individual-level values. \\

    Feasibility of requested DB/analytic method
    & 5 (11.9\%)
    & \textit{``Which country's microbial frequency is most similar to that of this ICU?''}
    & Country-level comparison data are unavailable in the database, requiring clarification of the intended comparison strategy or external reference source. \\

    \textit{Measurement source/data provenance}
    & 2 (4.8\%)
    & \textit{``What percentage of patients received a potassium chloride prescription (administration) within six hours of a reported potassium level of 3.5\,mEq/L?''}
    & Whether ``prescription'' refers to a medication order (\texttt{prescriptions}) or actual administration (\texttt{emar}). \\

    Anaphora/contextual reference
    & 2 (4.8\%)
    & \textit{``Test whether the association above is statistically significant.''}
    & Which previously discussed association is referenced by ``above.'' \\
\midrule
    \textbf{Total} & \textbf{42 (100\%)} & & \\
\bottomrule
    \end{tabular}
}
\end{table*}

To characterize why the conversational multi-turn interface requested clarification, each clarification exchange was assigned a single dominant trigger category based on the ambiguity or missing information that prompted the initial C-ANS response.
For exchanges involving multiple clarification rounds, only the initial C-ANS turn was coded because subsequent clarification requests were treated as part of the same exchange.
The taxonomy comprised six mutually exclusive categories, with three categories designated as requiring EHR- or clinical-domain knowledge.
Table~\ref{tab:typology_criteria} provides the definition and representative examples for each category.

The clarification-trigger typology showed that ambiguity arose from multiple sources rather than a single recurring cause.
Table~\ref{tab:typology_results} provides representative queries and the corresponding ambiguity underlying each trigger category.
The three categories most directly dependent on EHR- or clinical-domain knowledge---cohort/population definition, operational definition of a measure, and measurement source/data provenance---accounted for 27 of 42 clarification exchanges (64.3\%) and often could not be resolved without clinical knowledge or familiarity with the EHR schema.
Aggregation unit/grouping criterion was treated as a general ambiguity category because most cases involved domain-agnostic counting or grouping decisions. 
However, some cases, such as grouping by active ingredient or disease class, required pharmacological or ICD-classification knowledge, making this category partly domain-specific.
Other trigger categories reflected different interaction challenges.
Feasibility-related clarifications arose when the requested analysis could not be directly supported by the available database, while anaphora/contextual-reference clarifications arose when pronouns or other references were ambiguous with respect to earlier turns.

\subsection{Representative Qualitative Feedback on Interface Use}
\label{app:representative_feedback}
This section presents representative participant feedback supporting the qualitative observations summarized in Discussion section. 
Participant comments were translated from Korean into English for presentation, with ellipses indicating omitted portions of the original responses.

Several participants described progressive query refinement as a key advantage of the conversational multi-turn interface. 
P4 noted, \textit{``Users may not be able to formulate a clear question from the outset. They can make the question more specific as they ask questions and review the responses.''} 
P11 similarly emphasized that follow-up interaction could help users resolve ambiguity in their own queries: \textit{``Even when the initial question is ambiguous, follow-up questions allow users themselves to better understand the intended question and clarify its specific details.''} 
P2 further highlighted the practical relevance of this capability in clinical settings, noting that healthcare professionals may not always have the time or cognitive bandwidth to formulate highly specific prompts: \textit{``Healthcare professionals work under high workload and time pressure and often ask relatively abstract questions ... It is difficult to expect them to formulate every prompt as if they were writing a database query.''} 
These comments suggest that the conversational interface may reduce the burden of fully specifying an information need at the outset by allowing users to progressively refine it through dialogue.

Participants also identified the ability to revisit, verify, and correct results through continued dialogue as an important source of trust. 
P2 described a case in which responses across turns were inconsistent, but viewed the agent's subsequent recognition and correction of the discrepancy positively: \textit{``The earlier and later responses in the multi-turn interface were different, but the agent recognized the discrepancy itself, identified the reason, and presented it to the user ... which made me feel that I could move toward a more accurate final result.''} 
In contrast, P8 identified the inability to perform such verification as a limitation of the single-turn interface: \textit{``Because I could not ask additional questions, further verification required reformulating the question itself ... the interface felt less trustworthy because only a single question could be asked.''} 
These responses indicate that trust was influenced not only by the correctness of an initial answer, but also by whether users could interrogate, verify, and revise potentially uncertain outputs.
Not all participants, however, considered the conversational multi-turn interface more trustworthy. 
Four participants rated the two interfaces similarly in terms of trust. 
P9 pointed to limitations that could plausibly affect either interface, including oversimplification of clinical concepts and incomplete use of relevant variables: \textit{``When I asked the system to group disease categories, several different disease groups were simply combined into `others' ... and when I asked about patients' admission routes, it considered only `admission\_location' among several potentially relevant parameters ...''} 
This feedback indicates that limitations remain in the agent's representation of clinical concepts and selection of clinical keywords relevant to the user's intended analysis.

Finally, two participants indicated that the single-turn interface could still be useful for simple and well-defined tasks. 
P5 noted, \textit{``I could use the single-turn interface when asking a simple question ...''} 
Similarly, P9 commented, \textit{``It would be convenient when I want a short, direct answer to a simple question, so I would still use it ...''} 
These responses suggest that, despite the overall preference for the conversational multi-turn interface, the single-turn interface may remain useful for straightforward information retrieval when users already have a well-specified question and require only a concise answer.

\section{User Study Preparation}
\label{app:user_study}

\subsection{Agent Model Evaluation}
\label{app:agent_model_evaluation}

\subsubsection{Model Selection and Validation}
\label{app:agent_model_evaluation:model_selection}

\begin{table*}[t!]
    \caption{Comparison of GPT-5.2 and GPT-5.4 on the 45 Phase~1 fixed questions (FXQ; Table~\ref{tab:survey_fxq}). Each question was executed five times under the single-turn interface (225 trials per model). Accuracy was evaluated by comparing the generated SQL execution results with the corresponding gold result tables.
    }

    \label{tab:model_comparison}
    \resizebox{1\textwidth}{!}{
        \begin{tabular}{lrrrrrr}
        \toprule
        \textbf{Model} & \textbf{Avg. F1} & \textbf{Pass@1} & \textbf{Pass@5} & \textbf{Pass\textasciicircum 5} & \textbf{Avg. Latency (s)} & \textbf{Max Latency (s)} \\
        \midrule
        \textbf{GPT-5.2} & 65.4\% & 42.2\% & 55.6\% & 22.2\% & 5.74 & 10.86 \\
        \textbf{GPT-5.4} & \textbf{70.4\%} & \textbf{48.0\%} & \textbf{57.8}\% & \textbf{33.3\%} & \textbf{3.21} & \textbf{4.95} \\
        \bottomrule
        \end{tabular}
    }
    {\begin{itemize}
    \footnotesize
    \setlength{\itemsep}{0pt}
    \setlength{\parsep}{0pt}
    \setlength{\parskip}{0pt}
    \setlength{\topsep}{1pt}
    \setlength{\labelsep}{3pt}
    \renewcommand{\labelitemi}{\tiny$\bullet$}
    \item Avg.\ F1: Mean per-trial F1 score, computed using row-level matching with column-aware partial credit.
    \item Pass@1: Exact-match success rate of a single execution, representing the expected accuracy when only one attempt is allowed.
    \item Pass@5: Proportion of questions for which at least one of the five executions achieved an exact match.
    \item Pass\textasciicircum 5: Proportion of questions for which all five executions achieved an exact match.
    \item Avg./Max Latency: Mean and maximum response time from query submission to the final response, respectively.
    \end{itemize}
    }
\end{table*}

The first pilot study asked participants to interact with four candidate models (GPT-5 Nano~\citeapp{gpt5-app}, GPT-5.2~\citeapp{gpt5.2-app}, Gemini 3.0 Pro~\citeapp{gemini3.0-app}, and Qwen 3 32B~\citeapp{yang2025qwen3technicalreport-app}) so that the study designers could subsequently compare the accuracy of each model's responses to the same questions and select the model for the main study.
However, requiring participants to repeat the same tasks across four models substantially increased the experiment time and led to noticeable participant fatigue. 
Therefore, the second pilot study adopted a single-model protocol. 
Among the candidate models, GPT-5.2 demonstrated the most stable tool-calling behavior and the lowest response latency, and was therefore selected.

Before the main study began, GPT-5.4~\citeapp{gpt5.4-app} was released.
To determine the final model for the main study, we re-evaluated GPT-5.2 and GPT-5.4 under identical conditions using the 45 fixed questions (Table~\ref{tab:survey_fxq} in Appendix~\ref{app:survey_details}).
Each question was executed five times under the single-turn interface (225 trials per model), and the generated results were evaluated using multiple accuracy and latency metrics.
As shown in Table~\ref{tab:model_comparison}, GPT-5.4 outperformed GPT-5.2 on every accuracy metric while reducing average latency by 44\% and maximum latency by more than half.
Therefore, GPT-5.4 was adopted as the sole model for the main study.

\subsubsection{Conversational Performance Evaluation}
\label{app:agent_model_evaluation:conversational_eval}
\begin{table}[t!]
    \caption{Human-labeled correctness of GPT-5.4's conversational multi-turn responses, judged against the full preceding dialogue context, across all real Phase~2 sessions collected from the 14 main-study participants.}
    \label{tab:conversational_reliability}
    \centering
    \begin{tabular}{lrrr}
    \toprule
    \textbf{Category} & \textbf{Total Turns} & \textbf{Correct Turns} & \textbf{Accuracy} \\
    \midrule
    FXQ-turns               & 45  & 28  & 62.2\% \\
    Others (Q + SFQ + CLQ)  & 141 & 106 & 75.2\% \\
    \midrule
    All judgable turns      & 186 & 134  & 72.0\% \\
    \bottomrule
    \end{tabular}
\end{table}

Because the preceding benchmark evaluated only detailed, self-contained questions under the single-turn setting, it did not establish whether GPT-5.4 generated genuinely correct results during realistic conversational interactions, rather than merely producing plausible-looking responses. 
We therefore performed an additional evaluation after completion of the main study using all SQL-generating turns from the 57 Phase~2 conversational multi-turn sessions. 
For turns corresponding to the predefined fixed questions (FX-Q), correctness was assessed against the corresponding ground-truth tables.
For all other turns (Q + SF-Q + CL-Q), which did not have predefined ground truth, the study designers manually reviewed each generated result based on the full preceding dialogue context, including clarification and follow-up turns.
As shown in Table~\ref{tab:conversational_reliability}, 134 of the 186 evaluable turns were judged correct, yielding an overall accuracy of 72.0\%. 
Although participants expressed strong satisfaction with the conversational multi-turn interface in the user study (Fig.~\ref{fig:usability}), these results indicate that LLM-based database querying still has considerable room for improvement before routine real-world deployment.

To better understand the remaining errors, we reviewed the 52 incorrect turns and grouped them into recurring error categories.
The most common category, accounting for approximately one-third of the errors, involved keyword or entity matching. 
These errors arose from either overly broad substring matching (\eg matching \textit{epinephrine} within \textit{norepinephrine}) or overly narrow keyword lists that omitted clinically equivalent drug, diagnosis, or procedure terms.
Another common category involved population or aggregation mismatches, such as counting admissions or ICU stays instead of unique patients, or deduplicating records when the intended analysis required event-level counts.
A third category involved unrequested or overly restrictive conditions that substantially altered the target population.
For example, exact-timestamp joins between medication-order and administration records discarded the vast majority of otherwise valid matches, causing statistics to be computed from only a small fraction of the intended population.
The remaining errors resulted from misinterpreting database field semantics or representations, such as reversing the meaning of an administration-status flag or treating a date-only field as an exact event timestamp.

In addition to categorizing individual error types, we examined how errors propagated across conversational turns.
Approximately one-quarter of the incorrect turns reflected error propagation across the conversation.
Once an incorrect cohort definition, filtering condition, or aggregation rule had been established, subsequent turns often reused it without re-verification.
In several sessions, the agent also failed to preserve the dialogue state, redefining a previously established cohort or responding to a different follow-up option from the one selected by the participant.
These observations highlight a limitation specific to conversational multi-turn interactions: unlike errors in isolated single-turn queries, an early mistake or misinterpreted instruction can propagate through subsequent turns and affect multiple later results.

\subsection{User Study Interface}
\label{app:user_study_interface}
\begin{figure}[t!]
\centering
\includegraphics[width=0.8\textwidth]{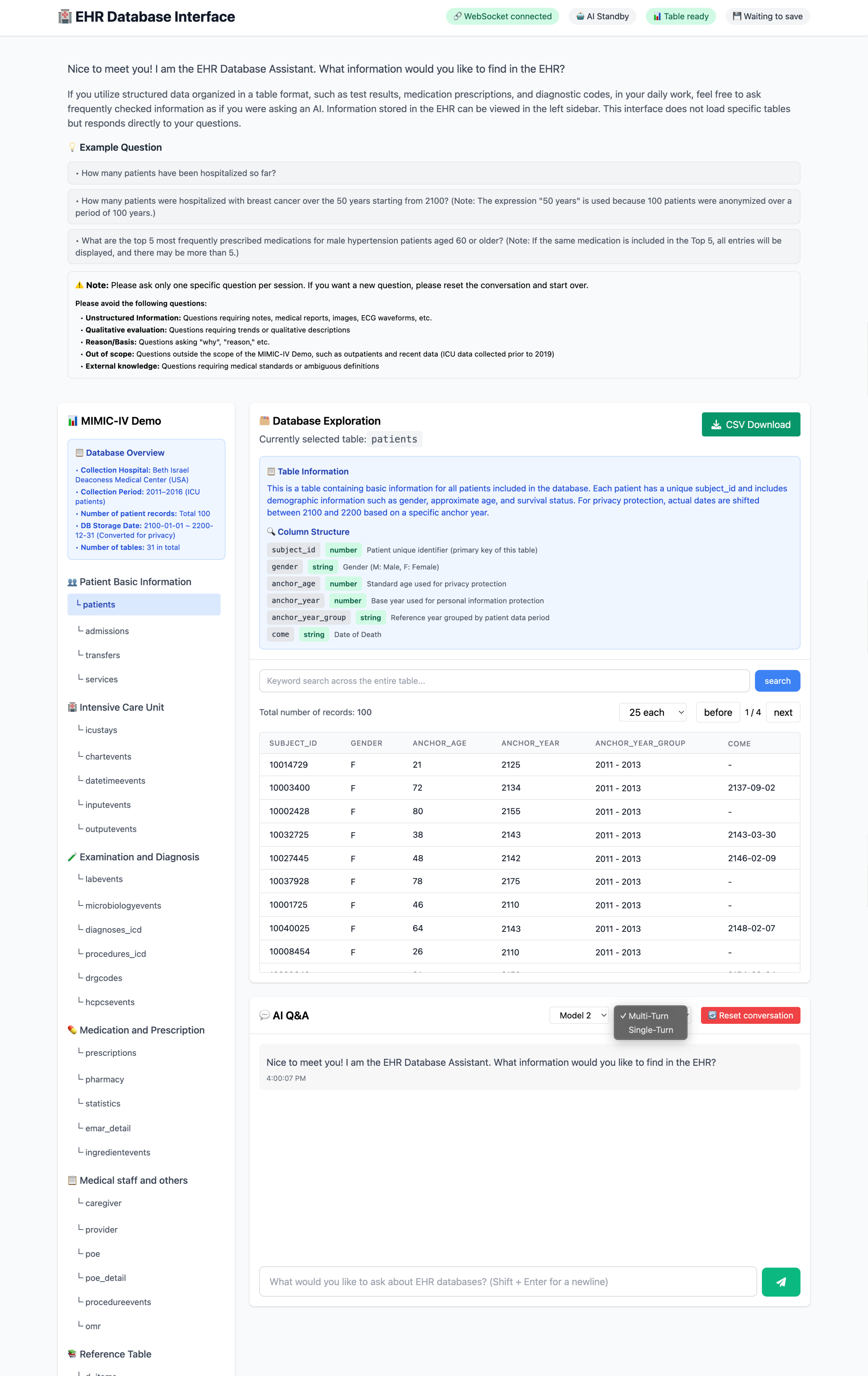}
    \caption{Web-based interface used in the user study. The left panel provides a schema browser listing database tables by category. The top-right panel displays the selected table's description, column-level schema, keyword search, and paginated raw data. The bottom-right panel provides the query interface, including model selection, interaction mode (single-turn or conversational multi-turn), and the conversation history.}
\label{fig:user_interface}
\end{figure}

Fig.~\ref{fig:user_interface} shows the web-based interface used in the user study. 
The interface integrated a schema browser, database explorer, and natural-language query interface into a single page to support both experimental conditions.

\noindent\textbf{Schema Browser.}
The left panel provided a categorized list of database tables together with basic dataset information. 
Selecting a table displayed its contents in the database explorer.

\noindent\textbf{Database Explorer.}
The upper-right panel displayed the selected table's description, column-level schema, keyword search, and paginated raw data to help participants understand the database structure.

\noindent\textbf{Query Interface.}
The lower-right panel provided the natural-language query interface. 
Participants could switch between the single-turn and conversational multi-turn modes, submit free-form questions, and view the conversation history. 
A model selector, session reset button, and tool-execution trace were also available.

\subsection{Survey Details}
\label{app:survey_details}
This section presents the full wording of all questionnaire items used in the user study.
Although the questionnaires were administered in Korean, English translations are provided here for reference.
The questionnaires consisted of six sections: Background Questionnaire, Phase 1 Evaluation, System Usability Scale (SUS), Unified Theory of Acceptance and Use of Technology (UTAUT), Query Detail Level (QDL), and Overall Comparison.
The SUS and UTAUT items were adapted to the interface evaluation context of this study while preserving the original intent of each item.
Table~\ref{tab:survey} summarizes all questionnaire items and their corresponding response formats, while Table~\ref{tab:survey_fxq} lists the 45 fixed questions (FX-Q) used in Phase 1, grouped according to whether answering each question required information from one, two, or three database tables.

\clearpage
{\footnotesize
    \begin{xltabular}{\linewidth}{c X >{\raggedleft\arraybackslash}p{2.4cm}}
        \caption{Questionnaire items and response formats used in the study (English translation; the original questionnaire was administered in Korean), grouped by questionnaire section.}
        \label{tab:survey} \\

        \toprule
        \textbf{No.} & \textbf{Item} & \textbf{Response Format} \\
        \midrule
        \endfirsthead

        \multicolumn{3}{c}{\tablename\ \thetable{} -- continued} \\
        \toprule
        \textbf{No.} & \textbf{Item} & \textbf{Response Format} \\
        \midrule
        \endhead

        \midrule
        \multicolumn{3}{r}{Continued on next page} \\
        \endfoot

        \bottomrule
        \endlastfoot

        \multicolumn{3}{l}{\textbf{Background Questionnaire}} \\
        \midrule
        1 & What is your current primary occupation? (\eg physician, nurse) & Open text \\
        2 & How many years of total clinical/professional experience do you have in the relevant field? & Categorical \\
        3 & Which department are you currently working in? & Open text \\
        4 & How frequently do you use EHR (or EMR) / CDW systems in your daily work? & Categorical \\
        5 & How would you rate your experience with statistical analysis tools (\eg SPSS, R, Python)? & 5-pt scale \\
        6 & How would you rate your prior knowledge of / experience with database query language (SQL)? & 5-pt scale \\
        7 & How frequently do you use conversational AI (\eg ChatGPT, Gemini) in your work or daily life? & Categorical \\

        \midrule

        \multicolumn{3}{l}{\textbf{Phase 1 Evaluation}} \\
        \midrule
        0-1 & Please list three MIMIC-IV-based questions that could also be useful in your actual clinical/professional work. & Open text \\
        0-2 & In what order did you complete the tasks? & Categorical \\
        1-1 & (Single-turn) Comments on your satisfaction with the answer to the question using 1 table. & Open text \\
        1-2 & (Multi-turn) Comments on your satisfaction with the answer to the question using 1 table. & Open text \\
        2-1 & (Single-turn) Comments on your satisfaction with the answer to the question using 2 tables. & Open text \\
        2-2 & (Multi-turn) Comments on your satisfaction with the answer to the question using 2 tables. & Open text \\
        3-1 & (Single-turn) Comments on your satisfaction with the answer to the question using 3 tables. & Open text \\
        3-2 & (Multi-turn) Comments on your satisfaction with the answer to the question using 3 tables. & Open text \\
        
        \midrule

        \multicolumn{3}{l}{\textbf{System Usability Scale (SUS)}} \\
        \midrule
        1-1 & (Single-turn) I would like to use this interface frequently. & 5-pt scale \\
        1-2 & (Multi-turn) I would like to use this interface frequently. & 5-pt scale \\
        2-1 & (Single-turn) I found this interface unnecessarily complex. (R) & 5-pt scale \\
        2-2 & (Multi-turn) I found this interface unnecessarily complex. (R) & 5-pt scale \\
        3-1 & (Single-turn) I found this interface easy to use. & 5-pt scale \\
        3-2 & (Multi-turn) I found this interface easy to use. & 5-pt scale \\
        4-1 & (Single-turn) I felt that I would need the support of a technical person to use this interface. (R) & 5-pt scale \\
        4-2 & (Multi-turn) I felt that I would need the support of a technical person to use this interface. (R) & 5-pt scale \\
        5-1 & (Single-turn) I found the various functions of this interface to be well integrated. & 5-pt scale \\
        5-2 & (Multi-turn) I found the various functions of this interface to be well integrated. & 5-pt scale \\
        6-1 & (Single-turn) I found this interface to be too inconsistent. (R) & 5-pt scale \\
        6-2 & (Multi-turn) I found this interface to be too inconsistent. (R) & 5-pt scale \\
        7-1 & (Single-turn) I believe that most people would learn to use this interface very quickly. & 5-pt scale \\
        7-2 & (Multi-turn) I believe that most people would learn to use this interface very quickly. & 5-pt scale \\
        8-1 & (Single-turn) I found this interface cumbersome to use. (R) & 5-pt scale \\
        8-2 & (Multi-turn) I found this interface cumbersome to use. (R) & 5-pt scale \\
        9-1 & (Single-turn) I felt confident using this interface. & 5-pt scale \\
        9-2 & (Multi-turn) I felt confident using this interface. & 5-pt scale \\
        10-1 & (Single-turn) I felt that I needed to learn many things before I could use this interface effectively. (R) & 5-pt scale \\
        10-2 & (Multi-turn) I felt that I needed to learn many things before I could use this interface effectively. (R) & 5-pt scale \\
        
        \midrule

        \multicolumn{3}{l}{\textbf{Unified Theory of Acceptance and Use of Technology (UTAUT)}} \\
        \midrule
        \multicolumn{3}{l}{\textbf{\quad UTAUT: Performance Expectancy (PE)}} \\
        \midrule
        1-1 & (Single-turn) I believe this interface would be useful in my work. & 5-pt scale \\
        1-2 & (Multi-turn) I believe this interface would be useful in my work. & 5-pt scale \\
        2-1 & (Single-turn) Using this interface would enable me to complete my work more quickly. & 5-pt scale \\
        2-2 & (Multi-turn) Using this interface would enable me to complete my work more quickly. & 5-pt scale \\
        3-1 & (Single-turn) Using this interface would improve my productivity. & 5-pt scale \\
        3-2 & (Multi-turn) Using this interface would improve my productivity. & 5-pt scale \\
        4-1 & (Single-turn) Using this interface would improve my job performance. & 5-pt scale \\
        4-2 & (Multi-turn) Using this interface would improve my job performance. & 5-pt scale \\
        \midrule
        \multicolumn{3}{l}{\textbf{\quad UTAUT: Effort Expectancy (EE)}} \\
        \midrule
        5-1 & (Single-turn) I found my interaction with this interface clear and understandable. & 5-pt scale \\
        5-2 & (Multi-turn) I found my interaction with this interface clear and understandable. & 5-pt scale \\
        6-1 & (Single-turn) I believe I could quickly become skillful at using this interface. & 5-pt scale \\
        6-2 & (Multi-turn) I believe I could quickly become skillful at using this interface. & 5-pt scale \\
        7-1 & (Single-turn) I found this interface easy to use. & 5-pt scale \\
        7-2 & (Multi-turn) I found this interface easy to use. & 5-pt scale \\
        8-1 & (Single-turn) I found it easy to learn how to use this interface. & 5-pt scale \\
        8-2 & (Multi-turn) I found it easy to learn how to use this interface. & 5-pt scale \\
        \midrule
        \multicolumn{3}{l}{\textbf{\quad UTAUT: Self-Efficacy (SE)}} \\
        \midrule
        9-1 & (Single-turn) I believe I could use this interface even if no one were available to help me. & 5-pt scale \\
        9-2 & (Multi-turn) I believe I could use this interface even if no one were available to help me. & 5-pt scale \\
        10-1 & (Single-turn) I believe I could complete the given task if someone were available to help when I encountered a problem. & 5-pt scale \\
        10-2 & (Multi-turn) I believe I could complete the given task if someone were available to help when I encountered a problem. & 5-pt scale \\
        11-1 & (Single-turn) Given enough time, I believe I could complete the given task using this interface. & 5-pt scale \\
        11-2 & (Multi-turn) Given enough time, I believe I could complete the given task using this interface. & 5-pt scale \\
        12-1 & (Single-turn) I believe I could complete the given task using only the interface's built-in help function. & 5-pt scale \\
        12-2 & (Multi-turn) I believe I could complete the given task using only the interface's built-in help function. & 5-pt scale \\
        \midrule
        \multicolumn{3}{l}{\textbf{\quad UTAUT: Anxiety (ANX)}} \\
        \midrule
        13-1 & (Single-turn) I felt anxious about using this interface. & 5-pt scale \\
        13-2 & (Multi-turn) I felt anxious about using this interface. & 5-pt scale \\
        14-1 & (Single-turn) I was afraid that I might lose important information by making a mistake while using this interface. & 5-pt scale \\
        14-2 & (Multi-turn) I was afraid that I might lose important information by making a mistake while using this interface. & 5-pt scale \\
        15-1 & (Single-turn) I hesitated to use this interface for fear of making a mistake that I could not undo. & 5-pt scale \\
        15-2 & (Multi-turn) I hesitated to use this interface for fear of making a mistake that I could not undo. & 5-pt scale \\
        16-1 & (Single-turn) I found this interface burdensome and intimidating to use. & 5-pt scale \\
        16-2 & (Multi-turn) I found this interface burdensome and intimidating to use. & 5-pt scale \\
        \midrule
        \multicolumn{3}{l}{\textbf{\quad UTAUT: Behavioral Intention (BI)}} \\
        \midrule
        17-1 & (Single-turn) I intend to use this interface in the future. & 5-pt scale \\
        17-2 & (Multi-turn) I intend to use this interface in the future. & 5-pt scale \\
        18-1 & (Single-turn) I expect to use this interface in the future (regardless of institutional policies or constraints). & 5-pt scale \\
        18-2 & (Multi-turn) I expect to use this interface in the future (regardless of institutional policies or constraints). & 5-pt scale \\
        19-1 & (Single-turn) I plan to use this interface in the future. & 5-pt scale \\
        19-2 & (Multi-turn) I plan to use this interface in the future. & 5-pt scale \\

        \midrule
        
        \multicolumn{3}{l}{\textbf{Query Detail Level (QDL) Scenarios}} \\
        \midrule
        1 & \textbf{Scenario 1:} Identifying a specific patient cohort. 
        \newline \textbf{[1] (Very general):} ``Show me the list of patients who've had high blood pressure recently.''
        \newline \textbf{[5] (Very detail):} ``Extract, in table form, the full list of ward patients whose systolic blood pressure was 140\,mmHg or higher over the past week.'' & 
        5-pt scale \\
        
        2 & 
        \textbf{Scenario 2:} Reviewing medication administration records.
        \newline \textbf{[1] (Very general):} ``What medications did patient A receive yesterday?'' 
        \newline \textbf{[5] (Very detail):} ``Show all oral and injectable medications administered to patient No.~12345 on Jan.~6, sorted by administration time.'' & 
        5-pt scale \\
        
        3 & 
        \textbf{Scenario 3:} Comparing laboratory test results.
        \newline \textbf{[1] (Very general):} ``Show how patient B's diabetes test results have changed.''
        \newline \textbf{[5] (Very detailed):} ``Compare patient B's HbA1c values over the past three months in chronological order and indicate any values exceeding the reference range.'' & 
        5-pt scale \\
        
        \midrule

        \multicolumn{3}{l}{\textbf{Overall Comparison}} \\
        \midrule
        1 & Considering your overall experience, which interface did you find more satisfying? & Categorical \\
        2 & If these interfaces were deployed in a real hospital database, which one would you prefer to use? & Categorical \\
        2-1 & What was the main reason for your choice? & Open text \\
        3 & Which interface did you find more trustworthy in terms of the information and results it provided? & Categorical \\
        3-1 & Please explain the reasons or experiences that influenced your judgment of trustworthiness. & Open text \\
        4 & Please share any suggestions for improving either interface or any additional comments about this study. & Open text \\
    \end{xltabular}
}

{\footnotesize
    \begin{xltabular}{\linewidth}{c X}
        \caption{The predefined question set (45 fixed questions) used in Phase 1 of the main study, grouped according to the number of database tables required to answer each question (English translation; the original questions were administered in Korean).}
        \label{tab:survey_fxq} \\

        \toprule
        \textbf{No.} & \textbf{Question} \\
        \midrule
        \endfirsthead

        \multicolumn{2}{c}{\tablename\ \thetable{} -- continued} \\
        \toprule
        \textbf{No.} & \textbf{Question} \\
        \midrule
        \endhead

        \midrule
        \multicolumn{2}{r}{Continued on next page} \\
        \endfoot

        \bottomrule
        \endlastfoot

        \multicolumn{2}{l}{\textbf{1-table Required}} \\
        \midrule
        1 & What is the average length of stay (in days) for discharged patients, and what are the longest and shortest lengths of stay? \\
        2 & How many ICU patients had a length of stay exceeding 7 days (long-stay patients), and what was their average length of stay? \\
        3 & What was the longest wait time among patients in the emergency department, and what is the average ED length of stay? \\
        4 & What is the sex distribution across all patients, and what are the average ages of male and female patients, respectively? \\
        5 & What is the distribution of admission routes among patients who died in the hospital, and how many deaths occurred via each route? \\
        6 & What are the top 10 most frequently prescribed medications, and how many prescriptions were issued for each? \\
        7 & What are the top 5 most frequently performed blood tests, and how many times was each performed? \\
        8 & What are the top 5 most frequently recorded vital sign items in ICU chart records? \\
        9 & What proportion of pharmacy-processed prescriptions were discontinued (status \texttt{`Discontinued'}), and how many prescriptions does that represent in total? \\
        10 & What is the distribution of specimen types with positive microbiology culture results, and which specimen type was cultured most frequently? \\
        11 & What are the in-hospital mortality rate (\%) and total number of admissions for inpatients, broken down by insurance type? \\
        12 & By ICU type (based on the first ICU ward of admission), what are the number of patients, total number of ICU admissions, and average length of stay (in days)? \\
        13 & When all patients are grouped into 10-year age brackets, what are the number of patients, number of deaths, and mortality rate for each group? \\
        14 & What are the top 10 organisms most frequently identified as antibiotic-resistant (\texttt{interpretation = `R'}) on microbiology culture, and how many patients were affected by each? \\
        15 & Among medications prescribed via the oral route (\texttt{route = `PO'}), what are the top 10 most frequently prescribed drugs, along with the number of prescriptions and number of unique patients for each? \\
        \addlinespace

        \midrule
        \multicolumn{2}{l}{\textbf{2-table Required}} \\
        \midrule
        
        1 & Please split patients into those whose ward was the same at ICU admission and discharge versus those whose ward changed, and compare the number of patients, average number of ward transfers, and average length of stay between the two groups. \\
        2 & For all patients, please calculate the average measured value of the \texttt{`Hemoglobin'} and \texttt{`Glucose'} lab test items, respectively. \\
        3 & During the same hospitalization, how many patients were transferred from the ICU to a general ward (or elsewhere) and subsequently readmitted to the ICU, and on average how many days elapsed between the first ICU discharge and ICU readmission? \\
        4 & Please check how prescribed \texttt{`Aspirin'} orders were actually processed (dispensing status) by the pharmacy, sorted by pharmacy ID, showing only the top 10 records. \\
        5 & For patient No.\ 10000032, among the electronic medication administration records that were not successfully completed, please list the drug name, route of administration, and reason for administration failure, showing only the 12 most recent records by administration time. \\
        6 & Among all patients with a history of ICU admission, please calculate the total number who ultimately died. \\
        7 & Among ICU patients, please extract the unique patient IDs of those with at least one record where a specific measurement item (item code: 220045) exceeded 100, sorted in ascending order of patient ID, listing only the first 16 patients without duplicates. \\
        8 & Please show every department and ward transfer that patient No.\ 10000032 went through from admission to discharge in chronological order. \\
        9 & For patients who died during hospitalization, please calculate the difference between admission time and discharge time for each, and derive the overall average total length of stay (in days). \\
        10 & Please report the average number of ward and department transfers, during hospitalization, for patients admitted via the emergency department. \\
        11 & By 10-year age bracket, what are the in-hospital mortality rate (\%) and average length of stay (in days) for inpatients? \\
        12 & What are the top 10 most frequently recorded primary diagnoses (\texttt{seq\_num = 1}), and how many patients are associated with each? \\
        13 & Among lab test items with at least 200 measurements, what are the top 10 blood test items with the highest proportion of abnormal (flag) results? \\
        14 & By ICU type, what is the proportion of oxygen saturation (SpO2, \texttt{itemid = 220277}) measurements below the normal range (95\%), and what is the average SpO2 value? \\
        15 & How do the proportion (\%) of patients prescribed the anticoagulant heparin and the total number of heparin prescriptions differ by sex? \\
        
        \midrule
        \multicolumn{2}{l}{\textbf{3-table Required}} \\
        \midrule
        
        1 & What is the total number of patients diagnosed with sepsis-related conditions, and what is their average ICU length of stay (in days)? \\
        2 & What are the top 5 most frequently performed procedures among patients admitted via the emergency department? \\
        3 & Among patients prescribed vancomycin, please report the sex distribution of those whose prescription remains active (not discontinued). \\
        4 & What is the average pH value for patients admitted to medical ICU-related wards? \\
        5 & Among the lab records of patients who died in the hospital, please retrieve the 20 most recent records with a lactate level of 4.0 or higher. \\
        6 & For patient No.\ 10000032's medication records, please show the list of drugs that were not actually administered along with their status at the time. \\
        7 & What is the average total length of stay (in days) for patients with \textit{Staphylococcus aureus} detected on microbiology culture? \\
        8 & Which departments were patients diagnosed with heart failure primarily assigned to or transferred through during hospitalization? \\
        9 & Among patients who were prescribed dopamine-class medications, how many survived to final discharge and how many died, respectively? \\
        10 & How do the average creatinine levels differ between patients aged 60 and older versus those younger than 60? \\
        11 & For patients diagnosed with pneumonia, what are the average length of stay (in days) and in-hospital mortality rate (\%), broken down by admission type? \\
        12 & By ICU type, what are the counts of fever (temperature above 100.4\textdegree{}F) and hypothermia (temperature below 96.8\textdegree{}F) readings, and the average temperature? \\
        13 & For emergency inpatients (\texttt{admission\_type LIKE `\%EMER\%'}), what are the top 10 most frequently performed procedures, along with the average length of stay and mortality rate for patients receiving each procedure? \\
        14 & Please compare the mean, minimum, and maximum serum creatinine values by sex and age group (under 50 / 50--64 / 65 and older). \\
        15 & By ICU type, what are the number of patients who received vasopressors (norepinephrine, vasopressin, epinephrine) and the total number of administration records? \\
    \end{xltabular}
}

\section{Quantitative Score Computation}
\label{app:sus_utaut}

\subsection{System Usability Scale (SUS)}
\label{app:sus}
Responses to the ten SUS items were converted into a composite score for each participant in both interface conditions following the standardized scoring procedure proposed by Brooke \etal~\citeapp{brooke1996sus-app}.
Positively worded items (1, 3, 5, 7, and 9) and negatively worded items (2, 4, 6, 8, and 10) were rescaled to a common 0--4 range after reverse-coding the negative items.
The rescaled scores were then summed and multiplied by 2.5, yielding a final score ranging from 0 to 100.

Let $X_i \in \{1,\dots,5\}$ denote a participant's raw Likert response to SUS item $i$ ($i=1,\dots,10$).
Items $i \in \{1,3,5,7,9\}$ are positively worded, and items $i \in \{2,4,6,8,10\}$ are negatively worded (reverse-coded).
The adjusted score $A_i$ for item $i$ is defined as

\begin{equation}
A_i =
\begin{cases}
X_i - 1, & i \in \{1,3,5,7,9\} \\
5 - X_i, & i \in \{2,4,6,8,10\}
\end{cases}
\label{eq:sus_adjusted}
\end{equation}

\noindent so that all items are aligned on a common scale, with 0 indicating the most negative and 4 the most positive evaluation.
The composite SUS score is then obtained by summing the ten adjusted item scores and rescaling to a 0--100 range:

\begin{equation}
\mathrm{SUS} = 2.5\sum_{i=1}^{10}A_i.
\label{eq:sus_total}
\end{equation}

\noindent This computation was performed independently for each participant's single-turn and conversational multi-turn sessions, yielding paired scores $\mathrm{SUS}_{\text{Single-turn}}$ and $\mathrm{SUS}_{\text{Multi-turn}}$.

\subsection{Unified Theory of Acceptance and Use of Technology (UTAUT)}
\label{app:utaut}
The 19 UTAUT items were grouped into five subscales (constructs) $c \in \{\mathrm{PE, EE, SE, ANX, BI}\}$: Performance Expectancy (PE, items 1--4), Effort Expectancy (EE, items 5--8), Self-Efficacy (SE, items 9--12), Anxiety (ANX, items 13--16), and Behavioral Intention (BI, items 17--19).
Let $I_c$ denote the set of items belonging to construct $c$, with $n_c = |I_c|$ ($n_{\mathrm{PE}} = n_{\mathrm{EE}} = n_{\mathrm{SE}} = n_{\mathrm{ANX}} = 4$, $n_{\mathrm{BI}} = 3$).
Each subscale score $S_c$ is computed as the simple mean of its constituent item responses $X_i \in \{1,\dots,5\}$:

\begin{equation}
S_c = \frac{1}{n_c}\sum_{i \in I_c} X_i.
\label{eq:utaut_subscale}
\end{equation}

\noindent Because Anxiety is scored in the opposite direction to the remaining four subscales, with higher values indicating greater apprehension rather than greater acceptance, it is reported separately.
An overall positive-acceptance score is computed as the mean of the four positively oriented subscales, excluding Anxiety:

\begin{equation}
\mathrm{Avg.\ UTAUT\ (P)} = \frac{S_{\mathrm{PE}} + S_{\mathrm{EE}} + S_{\mathrm{SE}} + S_{\mathrm{BI}}}{4}.
\label{eq:utaut_positive}
\end{equation}

\subsection{Statistical Analysis Details}
\label{app:statistical_analysis_details}
Because the study followed a within-subjects design, scores obtained from the same participant under the single-turn and conversational multi-turn conditions were analyzed as paired observations.
The same four-step procedure was applied independently to the SUS score, each of the five UTAUT constructs (PE, EE, SE, ANX, and BI), and the composite Avg.\ UTAUT (P) score.

First, a per-participant difference score was computed for each metric. 
Given the small sample size ($n=14$), the normality of the paired differences was assessed using the Shapiro-Wilk test~\citeapp{shaphiro1965analysis-app}. 
When the normality assumption was violated ($p_{\mathrm{SW}}<0.05$), the Wilcoxon signed-rank test~\citeapp{wilcoxon1945individual-app} was treated as the primary inferential test (Appendix~\ref{app:diff_normality}).
Second, a paired $t$-test was performed to obtain the test statistic ($t$), $p$-value, and Cohen's $d_z$~\citeapp{cohen2013statistical-app} as the measure of effect size (Appendix~\ref{app:paired_ttest}). 
Positive $t$ values indicate higher scores for the single-turn interface, and $|d_z|$ was interpreted using conventional thresholds of 0.2, 0.5, and 0.8 for small, medium, and large effects, respectively~\citeapp{brydges2019effect-app}.
Third, the Wilcoxon signed-rank test was performed for all metrics regardless of the normality outcome, reporting the test statistic ($W$), standardized statistic ($Z$), $p$-value, and effect size ($r$) (Appendix~\ref{app:wilcoxon}).
Positive $Z$ values indicate higher scores for the single-turn interface, and $|r|$ was interpreted using thresholds of 0.1, 0.3, and 0.5 for small, medium, and large effects, respectively~\citeapp{brydges2019effect-app}.
Finally, descriptive statistics and the results of the Shapiro--Wilk test, paired $t$-test, and Wilcoxon signed-rank test are reported for each metric.
Directional agreement between the paired $t$-test and Wilcoxon signed-rank test was considered when evaluating the consistency of the findings across statistical tests.

\subsubsection{Difference Scores and Normality Assessment}
\label{app:diff_normality}
For each paired metric (SUS, UTAUT subscales, and Avg.\ UTAUT (P)), a per-participant difference score was computed as

\begin{equation}
D_i = X_i^{S} - X_i^\text{M},
\label{eq:diff_score}
\end{equation}

\noindent where $X_i^{S}$ and $X_i^{M}$ denote participant $i$'s score under the single-turn and conversational multi-turn conditions, respectively.
Given the small sample size ($n=14$), the normality of $\{D_i\}$ was assessed with the Shapiro-Wilk test ($H_0$: the differences are drawn from a normal distribution; $p_{\mathrm{SW}}$ denotes the resulting $p$-value) prior to selecting the primary paired-comparison test:

\begin{equation}
T =
\begin{cases}
\text{Paired } t\text{-test}, & p_{\mathrm{SW}} \ge 0.05 \\
\text{Wilcoxon signed-rank test}, & p_{\mathrm{SW}} < 0.05
\end{cases}
\label{eq:test_selection}
\end{equation}

\subsubsection{Paired \textit{t}-test}
\label{app:paired_ttest}
The mean and standard deviation of the paired differences are

\begin{equation}
\bar{D} = \frac{1}{n}\sum_{i=1}^{n} D_i, \qquad
s_D = \sqrt{\frac{1}{n-1}\sum_{i=1}^{n}(D_i - \bar{D})^2},
\label{eq:diff_mean_sd}
\end{equation}

\noindent from which the paired $t$-test statistic and Cohen's effect size for paired samples are obtained as

\begin{equation}
t = \frac{\bar{D}}{s_D/\sqrt{n}}, \qquad df = n-1, \qquad d_z = \frac{\bar{D}}{s_D}.
\label{eq:paired_t}
\end{equation}

\subsubsection{Wilcoxon Signed-Rank Test}
\label{app:wilcoxon}
For the non-zero differences $D_i$, ranks $R_i$ were assigned to $|D_i|$ in ascending order (tied values receive the average rank), and the signed-rank sums were computed as

\begin{equation}
W^{+} = \sum_{D_i>0} R_i, \qquad W^{-} = \sum_{D_i<0} R_i, \qquad W = \min(W^{+}, W^{-}).
\label{eq:wilcoxon_w}
\end{equation}

\noindent For $m$ non-zero differences, the normal approximation to the sampling distribution of $W^{+}$ under $H_0$ has mean and standard deviation

\begin{equation}
\mu_W = \frac{m(m+1)}{4}, \qquad \sigma_W = \sqrt{\frac{m(m+1)(2m+1)}{24}},
\label{eq:wilcoxon_mu_sigma}
\end{equation}

\noindent yielding the standardized test statistic and the corresponding effect size:

\begin{equation}
Z = \frac{W^{+}-\mu_W}{\sigma_W}, \qquad r = \frac{Z}{\sqrt{m}}.
\label{eq:wilcoxon_z_r}
\end{equation}

\bibliographystyleapp{sn-nature}
\bibliographyapp{sn-bibliography}

\end{appendices}

\end{document}